\documentclass[journal]{IEEEtran}
\usepackage{cite}
\usepackage{amsmath,amssymb,amsfonts}
\usepackage{algorithmic}
\usepackage{graphicx}
\usepackage{textcomp}

\usepackage{bm}
\usepackage[T1]{fontenc}
\usepackage{hyperref, times, cite}
\usepackage{url}
\usepackage{enumitem}
\usepackage{mathtools}
\usepackage{bm, bbm}
\usepackage{cool}
\usepackage{amssymb,amsfonts,amsmath,amsthm,amscd,dsfont,mathrsfs}
\usepackage{graphicx,float,psfrag,epsfig,amssymb}
\usepackage{wrapfig}
\usepackage{relsize}
\usepackage{color}
\usepackage{pict2e}
\usepackage{algorithm}
\usepackage{caption}
\usepackage{nameref}
\usepackage{enumerate}
\usepackage{stackrel}

\usepackage{caption}
\usepackage{subcaption}

\usepackage{amsthm}

\begin{document}
\renewcommand{\algorithmicrequire}{\textbf{Input:}}
\renewcommand{\algorithmicensure}{\textbf{Output:}}
\thispagestyle{empty}

\title{Multi-Attribute Graph Estimation with Sparse-Group Non-Convex Penalties}
\author{ Jitendra K.\ Tugnait    
\thanks{J.K.\ Tugnait is with the Department of 
Electrical \& Computer Engineering,
200 Broun Hall, Auburn University, Auburn, AL 36849, USA. 
Email: tugnajk@auburn.edu . }

\thanks{This work was supported by the National Science Foundation Grant CCF-2308473.}}

\maketitle

\begin{abstract}
We consider the problem of inferring the conditional independence graph (CIG) of high-dimensional Gaussian vectors from multi-attribute data. Most existing methods for graph estimation are based on single-attribute models where one associates a scalar random variable with each node. In multi-attribute graphical models, each node represents a random vector. In this paper we provide a unified theoretical analysis of multi-attribute graph learning using a penalized log-likelihood objective function. We consider both convex (sparse-group lasso) and sparse-group non-convex (log-sum and smoothly clipped absolute deviation (SCAD) penalties) penalty/regularization functions. An alternating direction method of multipliers (ADMM) approach coupled with local linear approximation to non-convex penalties is presented for optimization of the objective function. For non-convex penalties, theoretical analysis establishing local consistency in support recovery, local convexity and precision matrix estimation in high-dimensional settings is provided under two sets of sufficient conditions: with and without some irrepresentability conditions. We illustrate our approaches using both synthetic and real-data numerical examples. In the synthetic data examples the sparse-group log-sum penalized objective function significantly outperformed the lasso penalized as well as SCAD penalized objective functions with $F_1$-score and Hamming distance as performance metrics.
\end{abstract}

\begin{IEEEkeywords}
Graph learning, inverse covariance estimation, undirected graph, sparse-group lasso, multi-attribute graphs, sparse-group log-sum and SCAD penalties.
\end{IEEEkeywords}

\section{Introduction} \label{intro}

\IEEEPARstart{G}{raphical} models provide a powerful tool for analyzing multivariate data \cite{Lauritzen1996, Whittaker1990}. In an undirected graphical model, the conditional dependency structure among $p$ random variables $x_1, x_2, \cdots , x_p$, (${\bm x} = [x_1 \; x_2 \; \cdots \; x_p]^\top$), is represented using an undirected graph ${\cal G} = \left( V, {\cal E} \right)$, where $V = \{1,2, \cdots , p\} =[p]$ is the set of $p$ nodes corresponding to the $p$ random variables $x_i$'s, and ${\cal E} \subseteq [p] \times [p]$ is the set of undirected edges describing conditional dependencies among $x_i$'s. The graph ${\cal G}$ is a conditional independence graph (CIG) where there is no edge between nodes $i$ and $j$  if and only if (iff) $x_i$ and $x_j$ are conditionally independent given the remaining $p$-$2$ variables. 

Gaussian graphical models (GGMs) are CIGs where ${\bm x}$ is multivariate Gaussian. Suppose ${\bm x}$ has positive-definite covariance matrix $\bm{\Sigma}$ with inverse covariance matrix $\bm{\Omega} = \bm{\Sigma}^{-1}$. Then ${\Omega}_{ij}$, the $(i,j)$-th element of  $\bm{\Omega}$, is zero iff $x_i$ and $x_j$ are conditionally independent. Given $n$ samples of ${\bm x}$, in high-dimensional settings, one estimates $\bm{\Omega}$ under some sparsity constraints; see, e.g., \cite{Buhlmann2011, Lam2009, Ravikumar2011, Wainwright2019}. In these graphs each node represents a scalar random variable. In many applications, there may be more than one random variable associated with a node. This class of graphical models has been called multi-attribute graphical models in \cite{Kolar2014, Tugnait21a, Tugnait2024} and vector graphs or networks in \cite{Marjanovic18, Sundaram20}. In \cite{Tugnait21a}, a sparse-group lasso \cite{Friedman2010a, Simon2013} based penalized log-likelihood approach for graph learning from multi-attribute data was presented whereas \cite{Kolar2014} considers only group lasso \cite{Yuan2006}.  Both sparse-group lasso and group lasso are convex penalties. It is well-known that use of non-convex penalties such as smoothly clipped absolute deviation (SCAD) \cite{Fan2001, Lam2009} or log-sum \cite{Candes2008}, can yield more accurate results. Such penalties can produce sparse set of solution like lasso, and approximately unbiased coefficients for large coefficients, unlike lasso. 

The objective of this paper is to investigate use of sparse-group non-convex log-sum and SCAD penalties for estimation of multi-attribute graphs. 

\subsection{Related Work}
Although non-convex penalties have been extensively used for graph estimation (see \cite{Lam2009, Zou2008, Loh2015, Loh2017, Tugnait21bb, Tugnait22} and references therein), its use for multi-attribute graph estimation is almost non-existent with the exception of \cite{Tugnait21b} where SCAD is investigated. 
In \cite{Tugnait21a}, a sparse-group lasso based penalized log-likelihood approach for graph learning from multi-attribute data was presented whereas  \cite{Kolar2014} considers only group lasso. In sparse-group lasso there is group level lasso penalty as well as element-wise lasso penalty (see equations (\ref{eqth2_20a}) and (\ref{eqth2_20b}) in Sec.\ \ref{PLL} in the sequel). We extend the approach of \cite{Tugnait21a} to non-convex sparse-group penalties. For analysis, we rely on the proof technique of \cite{Rothman2008} as well as the primal-dual witness technique of \cite{Ravikumar2011}, both originally used in the context of element-wise lasso penalty for single-attribute graphs. The technique of \cite{Ravikumar2011} was extended to group-lasso penalty for multi-attribute graphs in  \cite{Kolar2014}. The SCAD penalty for multi-attribute graphs has been considered in \cite{Tugnait21b} but it does not have counterparts to our Lemma 1 and Theorems 2-5. Moreover, the sparse-group SCAD penalty used in this paper is different than that in \cite{Tugnait21b}. In this paper we apply the primal-dual witness technique in a sparse-group non-convex penalty setting. Some prior results in an element-wise non-convex penalty setting are in \cite{Loh2015, Loh2017}.

For numerical optimization of the penalized log-likelihood we exploit an alternating direction method of multipliers (ADMM) approach \cite{Boyd2010} as in \cite{Tugnait21a, Tugnait21b}, where for non-convex penalties (SCAD or log-sum), we use a local linear approximation of the penalties (\cite{Zou2008, Lam2009, Tugnait21b}), initialized via sparse-group lasso results of \cite{Tugnait21a}.

\subsection{Our Contributions}
In this paper we provide a unified theoretical analysis of multi-attribute graph learning using a penalized log-likelihood objective function where both convex (sparse-group lasso) and non-convex (sparse-group log-sum and SCAD) penalty/regularization functions are considered. We establish  sufficient conditions in a high-dimensional setting for consistency (convergence of the precision matrix to true value in the Frobenius norm), local convexity when using non-convex penalties, and graph recovery. Two alternative sets of sufficient conditions are investigated, with and without some irrepresentability conditions. For the latter, we follow the proof technique of \cite{Rothman2008} (as in \cite{Tugnait21a}), and for the former, we follow the primal-dual witness technique of \cite{Ravikumar2011} (as in \cite{Kolar2014}), both applied in a sparse-group setting.  Theoretical results not relying on irrepresentability conditions are in Theorems 1-3 and that based on some irrepresentability conditions are in Theorems 4 and 5. While the non-convex penalized log-likelihood objective function results in  a non-convex optimization problem, Theorems 2 and 4 specify conditions under which it becomes a convex optimization problem (see Remark 2 in Sec.\ \ref{TA1}). These conditions favor log-sum penalty over SCAD.

A preliminary version of parts of this paper appears in a conference paper \cite{Tugnait24b}. Theorems 4-5, proof of Theorem 3, and synthetic and real data examples do not appear in \cite{Tugnait24b}. Only sketches of proofs of Theorems 1 and 2 appear in \cite{Tugnait24b}.

\subsection{Outline} \label{outnot}
The rest of the paper is organized as follows. The system model is presented in Sec.\ \ref{SM} where we describe the multi-attribute graphical model with $m$ random variables per node, and also an associated larger single-attribute graph. A penalized log-likelihood objective function is presented in Sec.\ \ref{PLL} for estimation of multi-attribute graph using non-convex penalties. An ADMM approach coupled with a local linear approximation to non-convex penalties is presented for optimization of the objective function in Sec.\ \ref{OPT}. In Sec.\ \ref{TA} we present sufficient conditions in a high-dimensional setting for consistency, local convexity when using non-convex penalties, and graph recovery. Numerical results based on synthetic as well as real data are presented in Sec.\ \ref{NE} to illustrate the proposed approach.  Proofs of Theorems 1-5 and related technical lemmas are given in Appendices \ref{append1}, \ref{append2}  and \ref{append3}.

\subsection{Notation} \label{outnot2}
For a set $V$, $|V|$ or $\mbox{card}(V)$ denotes its cardinality. The abbreviations w.p.\ and w.h.p.\ stand for with probability and with high probability, respectively. Given ${\bm A} \in \mathbb{R}^{p \times p}$, we use $\phi_{\min }({\bm A})$, $\phi_{\max }({\bm A})$, $|{\bm A}|$ and $\mbox{tr}({\bm A})$ to denote the minimum eigenvalue, maximum eigenvalue, determinant and  trace of ${\bm A}$, respectively. For ${\bm B} \in \mathbb{R}^{p \times q}$, we define  $\|{\bm B}\| = \sqrt{\phi_{\max }({\bm B}^\top  {\bm B})}$, $\|{\bm B}\|_F = \sqrt{\mbox{tr}({\bm B}^\top  {\bm B})}$, $\|{\bm B}\|_1 = \sum_{i,j} |B_{ij}|$, where $B_{ij}$ is the $(i,j)$-th element of ${\bm B}$ (also denoted by $[{\bm B}]_{ij}$), $\|{\bm B}\|_\infty = \max_{i,j} |B_{ij}|$ and $\|{\bm B}\|_{1,\infty} = \max_i \sum_j|B_{ij}|$. Given ${\bm A} \in \mathbb{R}^{p \times p}$, ${\bm A}^+ = \mbox{diag}({\bm A})$ is a diagonal matrix with the same diagonal as ${\bm A}$, and  ${\bm A}^- = {\bm A} - {\bm A}^+$ is ${\bm A}$ with all its diagonal elements set to zero. The notation ${\bm y}_n = {\cal O}_P({\bm x}_n)$ for random vectors ${\bm y}_n, {\bm x}_n \in \mathbb{R}^p$ means that for any $\varepsilon > 0$, there exists $0 < M < \infty$ such that $P ( \|{\bm y}_n\| \le M \|{\bm x}_n\|) \ge 1 - \varepsilon$ $\forall n \ge 1$.
The symbols $\otimes$ and $\boxtimes$ denote Kronecker product and Tracy-Singh product \cite{Tracy1989}, respectively. In particular, given block partitioned matrices ${\bm A} =[{\bm A}_{ij}]$ and ${\bm B}=[{\bm B}_{k \ell}]$  with submatrices ${\bm A}_{ij}$ and ${\bm B}_{k \ell}$, Tracy-Singh product yields another block partitioned matrix ${\bm A} \boxtimes {\bm B} = [{\bm A}_{ij} \boxtimes {\bm B}]_{ij} = [[{\bm A}_{ij} \otimes {\bm B}_{k \ell}]_{k \ell} ]_{ij}$ \cite{Liu2008}. Given ${\bm A} =[{\bm A}_{ij}] \in \mathbb{R}^{mp \times mp}$ with ${\bm A}_{ij} \in \mathbb{R}^{m \times m}$, ${\rm vec}({\bm A}) \in \mathbb{R}^{m^2p^2}$ denotes the vectorization of ${\bm A}$ which stacks the columns of the matrix ${\bm A}$, and 
\begin{align*}
  & {\rm bvec}({\bm A}) =  [({\rm vec}({\bm A}_{11}))^\top \; ({\rm vec}({\bm A}_{21}))^\top \; \cdots 
      \; ({\rm vec}({\bm A}_{p1}))^\top \\
		& \quad ({\rm vec}({\bm A}_{12}))^\top \;  \cdots \; ({\rm vec}({\bm A}_{p2}))^\top \; \cdots \; ({\rm vec}({\bm A}_{pp}))^\top]^\top .
\end{align*}

Let $V = [p]$, $T \subseteq V \times V$, ${\bm A} \in \mathbb{R}^{mp \times mp}$ and let ${\bm A}^{(k \ell)} \in \mathbb{R}^{m \times m}$ denote an $m \times m$ submatrix of ${\bm A}$ with $k$ and $\ell$ indexing some $m$ rows and $m$ columns, respectively, of ${\bm A}$. Then ${\bm A}_T$ denotes the submatrix of ${\bm A}$ with rows and columns indexed by $T$, i.e., ${\bm A}_T =[{\bm A}^{(k \ell)}]_{(k , \ell)  \in T}$. Suppose ${\bm \Gamma} = {\bm A} \boxtimes {\bm B}$ given block partitioned matrices ${\bm A} =[{\bm A}_{ij}]$ and ${\bm B}=[{\bm B}_{k \ell}]$. For any two subsets  $T_1$ and $T_2$ of  $V \times V$, ${\bm \Gamma}_{T_1,T_2}$ denotes the submatrix of ${\bm \Gamma}$ with block rows and columns indexed by $T_1$ and $T_2$, i.e., ${\bm \Gamma}_{T_1,T_2} = [{\bm A}_{j \ell} \otimes {\bm B}_{kq}]_{(j,k) \in T_1, (\ell,q) \in T_2}$. Following \cite{Kolar2014}, an operator $\bm{\mathcal C}( \cdot )$ is used in Sec.\ \ref{TA2}. Consider ${\bm A} \in \mathbb{R}^{mp \times mp}$ with $(k,l)$th $m \times m$ submatrix ${\bm A}^{(k \ell)}$. Then $\bm{\mathcal C}( \cdot )$ operates on ${\bm A}$ as 
\begin{align*}
 \begin{bmatrix} {\bm A}^{(11)} & \cdots & {\bm A}^{(1p)} \\
			  \vdots     & \ddots     & \vdots \\
			{\bm A}^{(p1)} &  \cdots & {\bm A}^{(pp)} \end{bmatrix} \overset{\bm{\mathcal C}( \cdot )}{\xrightarrow{\hspace*{0.5cm}}}
			\begin{bmatrix} \|{\bm A}^{(11)}\|_F & \cdots & \|{\bm A}^{(1p)}\|_F \\
			  \vdots     & \ddots     & \vdots \\
			\|{\bm A}^{(p1)}\|_F &  \cdots & \|{\bm A}^{(pp)}\|_F \end{bmatrix}
\end{align*}
with $\bm{\mathcal C}( {\bm A}^{(k \ell)} ) = \|{\bm A}^{(k \ell)}\|_F$ and $\bm{\mathcal C}( {\bm A} ) \in \mathbb{R}^{p \times p}$. Now consider ${\bm A}_1  , {\bm A}_2 \in \mathbb{R}^{mp \times mp}$ with $(k,l)$th $m \times m$ submatrices ${\bm A}_1^{(k \ell)}$ and ${\bm A}_2^{(k \ell)}$, respectively, and Tracy-Singh product ${\bm A}_1 \boxtimes {\bm A}_2 \in \mathbb{R}^{(mp)^2 \times (mp)^2}$. Then $\bm{\mathcal C}( \cdot )$ operates on ${\bm A}_1 \boxtimes {\bm A}_2$ as $\bm{\mathcal C}({\bm A}_1 \boxtimes {\bm A}_2) \in \mathbb{R}^{p^2 \times p^2}$ with $\bm{\mathcal C}( {\bm A}_1^{(k_1 \ell_1)} \otimes {\bm A}_2^{(k_2 \ell_2)} ) = \|{\bm A}_1^{(k_1 \ell_1)} \otimes {\bm A}_2^{(k_2 \ell_2)} \|_F$ (=$\|{\bm A}_1^{(k_1 \ell_1)}  \|_F \, \|{\bm A}_2^{(k_2 \ell_2)} \|_F$). That is, each $m^2 \times m^2$ submatrix ${\bm A}_1^{(k_1 \ell_1)} \otimes {\bm A}_2^{(k_2 \ell_2)}$ of ${\bm A}_1 \boxtimes {\bm A}_2 $ is mapped into its Frobenius norm.

\section{System Model} \label{SM}
We will call ${\cal G}$ considered earlier a {\it single-attribute graphical model} for ${\bm x}$. Now consider $p$ jointly Gaussian random vectors ${\bm z}_i \in \mathbb{R}^m$, $i \in [p]$. We associate ${\bm z}_i$ with the $i$th node of an undirected graph ${\cal G} = (V, {\cal E})$ where $V= [p]$ and edges in ${\cal E}$ describe the conditional dependencies among vectors $\{ {\bm z}_i, \; i \in V\}$. As in the scalar case ($m=1$), there is no edge between node $i$ and node $j$ in ${\cal G}$ iff random vectors ${\bm z}_i$ and ${\bm z}_j$ are conditionally independent given all the remaining random vectors \cite{Kolar2014}. This is the {\it multi-attribute Gaussian graphical model} of interest in this paper. 

Define the $mp$-vector
\begin{equation} \label{addn010}
   {\bm x} = [ {\bm z}_1^\top \; {\bm z}_2^\top \; \cdots \; {\bm z}_p^\top ]^\top \in \mathbb{R}^{mp} \, .
\end{equation}
Suppose we have $n$ i.i.d.\ observations ${\bm x}(t)$, $t=1,2, \cdots , n$, of zero-mean ${\bm x}$. Our objective is to estimate the inverse covariance matrix $(\mathbb{E}\{{\bf x}{\bf x}^\top\})^{-1}$ and to determine if edge $\{ i,j \} \in {\cal E}$, given data $\{ {\bm x}(t) \}_{t=1}^{n}$.
Let us associate ${\bm x}$ with an ``enlarged'' graph $\bar{\cal G} = (\bar{V}, \bar{\cal E})$, where $\bar{V} = [mp]$ and $\bar{\cal E} \subseteq \bar{V} \times \bar{V}$. Now $[{\bm z}_j]_\ell$, the $\ell$th component of ${\bm z}_j$ associated with node $j$ of ${\cal G} = (V, {\cal E})$, is the random variable $x_q = [{\bm x}]_q$, where $q = (j-1)m+\ell$, $j \in [p]$ and $\ell \in [m]$. The random variable $x_q$ is associated with node $q$ of $\bar{\cal G} = (\bar{V}, \bar{\cal E})$. Corresponding to the edge $\{j,k\} \in {\cal E}$ in the multi-attribute ${\cal G} = (V, {\cal E})$, there are $m^2$ edges $\{ q,r\} \in \bar{\cal E}$ specified by $q=(j-1)m+s$ and $r=(k-1)m+t$, where $s,t \in [m]$. The graph $\bar{\cal G} = (\bar{V}, \bar{\cal E})$ is a single-attribute graph. In order for $\bar{\cal G}$ to reflect the conditional independencies encoded in ${\cal G}$, we must have the equivalence
\[ 
  \{ j,k \}  \not\in {\cal E} \, \Leftrightarrow  \,  
	  \bar{\cal E}^{(jk)} \cap \bar{\cal E} =\emptyset 
\] 
where 
\begin{align*}
\bar{\cal E}^{(jk)} = &  \big\{ \{q,r\} \, : \,  q=(j-1)m+s, \, r=(k-1)m+t, \\
	& \hspace*{0.5in}  s,t \in [m] \big\} .
\end{align*}
Let ${\bm R}_{xx} = \mathbb{E}\{ {\bm x} {\bm x}^\top \} \succ {\bm 0}$ and $\bm{\Omega} = {\bm R}_{xx}^{-1}$. Define the $(j,k)$th $m \times m$ subblock $\bm{\Omega}^{(jk)}$ of $\bm{\Omega}$ as
\begin{equation} \label{addn020}
   [\bm{\Omega}^{(jk)}]_{st} = [ \bm{\Omega} ]_{(j-1)m+s,(k-1)m+t} \, , \; s,t \in [m] \, .
\end{equation}
It is established in \cite[Sec.\ 2.1]{Kolar2014} that 
$\bm{\Omega}^{(jk)} = {\bm 0}  \Leftrightarrow \, \{ j,k \} \not\in {\cal E}$.
Since $\bm{\Omega}^{(jk)} = {\bm 0}$ is equivalent to $[\bm{\Omega}]_{qr} = 0$ for every $\{ q,r\} \in \bar{\cal E}^{(jk)}$, and since, by \cite[Proposition 5.2]{Lauritzen1996},  $[\bm{\Omega}]_{qr} = 0$ iff $x_q$ and $x_r$ are conditionally independent, hence, iff $\{q,r\} \not\in \bar{\cal E}$, it follows that the aforementioned equivalence holds true.

\section{Penalized Negative Log-Likelihood}  \label{PLL}
Consider a finite set of data comprised of $n$ i.i.d.\ zero-mean observations  ${\bm x}(t)$, $t=1,2, \cdots , n$. Parametrizing in terms of the precision (inverse covariance) matrix $\bm{\Omega}$, the negative log-likelihood, up to some irrelevant constants, is given by
\begin{align} 
    {\cal L}({\bm \Omega}) 
					&	:=  -\ln (|\bm{\Omega}|) 
		        + {\rm tr} \left( \hat{\bm \Sigma} \bm{\Omega} \right)     	 \label{eqth2_10}
\end{align}
where
\begin{equation}  \label{moresm} 
	 \hat{\bm \Sigma} =  \frac{1}{n} \sum_{t=1}^{n} {\bm x}(t) {\bm x}^\top(t) \, . 
\end{equation}

In the high-dimensional case ($n < p$ or $n$ comparable to $p$), to enforce sparsity and to make the problem well-conditioned, we propose to minimize a penalized version  $\bar{\cal L}({\bm \Omega})$ of ${\cal L}({\bm \Omega})$ where we penalize (regularize) both element-wise and group-wise.  We have
\begin{align}  
   \bar{\cal L}({\bm \Omega}) & =  {\cal L}({\bm \Omega}) + \alpha P_e({\bm \Omega}) 
	   + (1-\alpha) P_g({\bm \Omega}) , \label{eqth2_20} 
\end{align}
\begin{align}
	   P_e({\bm \Omega}) &  =   \sum_{i \ne j}^{mp} 
		  \rho_\lambda \left( \big| [ {\bm{\Omega}} ]_{ij} \big| \right)  , \label{eqth2_20a} 
\end{align}
\begin{align}
		P_g({\bm \Omega}) &  = m  \, \sum_{ q \ne \ell}^p \; 
		    \rho_\lambda \left( \| {\bm \Omega}^{(q \ell)} \|_F \right) \label{eqth2_20b} 
\end{align}
where ${\bm \Omega}^{(q \ell )} \in \mathbb{R}^{m \times m}$ is defined as in (\ref{addn020}), $\lambda > 0$, $\alpha \in [0,1]$, $m $ in (\ref{eqth2_20b}) reflects the number of group variables \cite{Yuan2006}, and for $u \in \mathbb{R}$, $\rho_\lambda(u)$ is a penalty function that is function of $|u|$. In (\ref{eqth2_20a}), the penalty term is applied to each off-diagonal element of $\bm{\Omega}$ and in (\ref{eqth2_20b}), the penalty term is applied to the off-block-diagonal group of $m^2$ terms via ${\bm \Omega}^{(q \ell)}$. The parameter $\alpha \in [0,1]$ ``balances'' element-wise and group-wise penalties \cite{Friedman2010a, Simon2013, Tugnait21a}.

The following penalty functions are considered:
\begin{itemize}
\item {\it Lasso}. For some $\lambda > 0$,  
       $ \rho_\lambda(u) = \lambda |u| , \quad u \in \mathbb{R}$.
\item {\it Log-sum}. For some $\lambda > 0$ and $1 \gg \epsilon > 0$, 
        $\rho_\lambda(u) = \lambda \epsilon \, \ln \left( 1 + \frac{|u|}{\epsilon} \right)$.
\item {\it SCAD}. For some $\lambda > 0$ and $a > 2$, 
\begin{equation}  \label{scad}
   \rho_\lambda(u) = \left\{ \begin{array}{ll}
	   \lambda | u | \, , &  |u| \le \lambda,  \\
		\frac{2 a \lambda | u |- | u |^2 - \lambda^2}{2 (a-1)}  \, ,
		      & \lambda < |u| < a \lambda \\
		\frac{\lambda^2 (a+1)}{2} \, , & |u| \ge a \lambda . \end{array} \right.
\end{equation}
\end{itemize}
In the terminology of \cite{Loh2017}, all of the above three penalties are ``$\mu$-amenable'' for some $\mu \ge 0$. As defined in \cite[Sec.\ 2.2]{Loh2017}, $\rho_\lambda(u)$ is $\mu$-amenable for some $\mu \ge 0$ if 
\begin{itemize}
\item[(i)] The function $\rho_\lambda(u)$ is symmetric around zero, i.e., $\rho_\lambda(u) = \rho_\lambda(-u)$ and $\rho_\lambda(0) = 0$.
\item[(ii)] The function $\rho_\lambda(u)$ is nondecreasing on $\mathbb{R}_+$.
\item[(iii)] The function $\rho_\lambda(u)/u$ is nonincreasing on $\mathbb{R}_+$.
\item[(iv)] The function $\rho_\lambda(u)$ is differentiable for $u \ne 0$.
\item[(v)] The function $\rho_\lambda(u) +\frac{\mu}{2} u^2$ is convex, for some $\mu \ge 0$.
\item[(vi)] $\lim_{u \rightarrow 0^+} \frac{d \rho_\lambda (u)}{du} = \lambda$.
\end{itemize}
It is shown in \cite[Appendix A.1]{Loh2017}, that all of the above three penalties are $\mu$-amenable with $\mu = 0$ for Lasso and $\mu =1/(a-1)$ for SCAD. In \cite{Loh2017} the log-sum penalty is defined as $\rho_\lambda(u) = \ln (1+\lambda |u|)$ whereas in \cite{Candes2008}, it is defined as $\rho_\lambda(u) = \lambda \, \ln \left( 1 + \frac{|u|}{\epsilon} \right)$. We follow  \cite{Candes2008} but modify it so that property (vi) in the definition of $\mu$-amenable penalties holds. In our case $\mu = \frac{\lambda}{\epsilon}$ for the log-sum penalty since $\frac{d^2 \rho_\lambda (u)}{du^2} = - \lambda \epsilon /(\epsilon + |u|)^2$ for $u \ne 0$.

The  following properties also hold for the three penalty functions:
\begin{itemize}
\item[(vii)] For some $C_\lambda > 0$ and $\delta_\lambda > 0$, we have 
\begin{equation}  \label{prop7} 
        \rho_\lambda(u) \ge C_\lambda |u| \mbox{  for  } |u| \le \delta_\lambda \, .
\end{equation}
\item[(viii)] $\frac{d \rho_\lambda (u)}{d |u|} \le \lambda$ for $u \ne 0$.
\end{itemize}
Property (viii) is straightforward to verify. For Lasso,  $C_\lambda = \lambda$ and $\delta_\lambda = \infty$. For SCAD, $C_\lambda = \lambda$ and $\delta_\lambda = \lambda$. Since $\ln (1+x) \ge x/(1+x)$ for $x > -1$, we have $\ln (1+x) \ge x/C_1$ for $0 \le x \le C_1-1$, $C_1 > 1$. Take $C_1 =2$. Then log-sum $\rho_\lambda(u) \ge \frac{\lambda}{2} |u|$ for any $|u| \le \epsilon$, leading to $C_\lambda = \frac{\lambda}{2}$ and $\delta_\lambda =\epsilon$. We may and will take $C_\lambda = \frac{\lambda}{2}$ for lasso and SCAD penalties as well.

We seek $\hat{\bm \Omega} = \arg\min_{{\bm \Omega} \succ {\bm 0}} \bar{\cal L}({\bm \Omega})$.

\section{Optimization} \label{OPT}
The objective function $\bar{\cal L}({\bm \Omega})$ is  non-convex for the non-convex SCAD and log-sum penalties. In this section we discuss an ADMM approach, following the ADMM approach given in \cite{Tugnait21a} for sparse group lasso, to attain a local minimum of $\bar{\cal L}({\bm \Omega})$ w.r.t.\ ${\bm \Omega}$.  

For non-convex $\rho_\lambda(u)$, we use a local linear approximation (LLA) to $\rho_\lambda(u)$ as in \cite{Zou2008, Lam2009}, to yield 
\begin{equation}
  \rho_{\lambda}(u) \approx \rho_{\lambda}(|u_0|) 
	 + \rho_\lambda^\prime(|u_0|) (|u| - |u_0|) \, ,
\end{equation}
where $u_0$ is an initial guess, and the gradient of the penalty function is 
\begin{align} 
  \rho_\lambda^\prime(|u_0|) = & \left\{ \begin{array}{l}
							\frac{\lambda \epsilon }{|u_0| +\epsilon}   \mbox{ for log-sum}, \\ 
			 \left\{ \begin{array}{ll} \lambda , & \mbox{if  } |u_0| \le \lambda \\
				    \frac{ a \lambda - | u_0 |}{ a-1} , 
									& \mbox{if  } \lambda < |u_0| \le a \lambda \\ 
						0 , & \mbox{if  } a \lambda < |u_0|  \end{array} \right. \\
						  \quad\quad \mbox{ for SCAD}. \end{array} \right.  
\end{align}
Therefore, with $u_0$ fixed, we need to consider only the term dependent upon $u$ for optimization w.r.t.\ $u \,$:
\begin{equation}
  \rho_{\lambda}(u)  \, \Rightarrow \, \rho_\lambda^\prime(|u_0|) \, |u| \, .
\end{equation}
By \cite[Theorem 1]{Zou2008}, the LLA provides a majorization of the non-convex penalty, thereby yielding a majorization-minimization approach. By \cite[Theorem 2]{Zou2008}, the LLA is the best convex majorization of the LSP and SCAD penalties.

\begin{algorithm} 
\caption{ADMM Algorithm for Sparse-Group Graphical Lasso}
\label{alg0}

\algorithmicrequire{\; Sample covariance $\hat{\bm{\Sigma}}$ (see (\ref{moresm})), regularization and penalty parameters $\lambda_{e,ij}$ ($i,j \in [mp]$), $\lambda_{g, k \ell}$ ($k, \ell \in [p]$), $\alpha$ and $\rho=\bar{\rho}$, tolerances $\tau_{abs}$ and $\tau_{rel}$, variable penalty factor $\phi$, maximum number of iterations $t_{max}$}. Initial guess $\bar{\bm \Omega}$. \\
\algorithmicensure{\;\ estimated inverse covariance $\hat{\bm \Omega}$ and edge-set $\hat{\cal E}$}

\begin{algorithmic}[1] 
\STATE Initialize: ${\bm U}^{(0)} = {\bm V}^{(0)} = {\bm 0}$, ${\bm \Omega}^{(0)} = \bar{\bm \Omega}$, where ${\bm U}, {\bm V} \in \mathbb{R}^{(mp) \times (mp)}$, $\rho^{(0)} = \bar{\rho}$ 
\STATE converged = \FALSE, $t=0$
\WHILE{converged = \FALSE $\;$ \AND $\;$ $t \le t_{max}$,}
\STATE Eigen-decompose $\hat{\bm \Sigma} - \rho^{(t)} \left( {\bm V}^{(t)} - {\bm U}^{(t)} \right)$ as $\hat{\bm \Sigma} - \rho^{(t)} \left( {\bm V}^{(t)} - {\bm U}^{(t)} \right) = {\bm P}{\bm D}{\bm P}^\top$ with diagonal matrix ${\bm D}$ consisting of eigenvalues.  Define diagonal matrix $\tilde{\bm D}$ with $\ell$th diagonal element
$\tilde{\bm D}_{\ell \ell} = ( -{\bm D}_{\ell \ell} + \sqrt{ {\bm D}_{\ell \ell}^2 + 4 \rho^{(t)}  } \, )/(2 \rho^{(t)})$. Set $\bm{\Omega}^{(t+1)} = {\bm P} \tilde{\bm D} {\bm P}^\top$.
\STATE Define soft thresholding scalar operator $T_{st}(a, \beta) := (1-\beta/|a|)_+ a$. Set ${\bm A}^{(k \ell)} = (\bm{\Omega}^{(t+1)})^{(k \ell)} + ({\bm U}^{(t)})^{(k \ell)}$. The diagonal $m \times m$ subblocks of ${\bm V}$ are updated as 
\begin{align*}  
    [({\bm V}^{(t+1)})^{(kk)}]_{u v} & = 
	 \left\{ \begin{array}{l}
			    [\bm{A}^{(kk)}]_{uu}  \quad \mbox{  if } u=v\\
					\hspace*{-0.1in} T_{st}([{\bm A}_k^{(kk)}]_{uv}, \frac{ \alpha \lambda_{e,ij}}{\rho^{(t)}}) 
					   \,  \mbox{ if } u \neq v \end{array} \right.
\end{align*}
$k \in [p]$, $\; u,v \in [m]$, $i=(k-1)m+u$, $j=(k-1)m+v$. The off-diagonal $m \times m$ subblocks of ${\bm V}$ are updated as 
\begin{align*}  
   ({\bm V}^{(t+1)})^{(k \ell)} & = 
	 {\bm B} \Big( 1 - \frac{(1-\alpha) m \lambda_{g,k \ell}}
					    {\rho^{(t)} \| {\bm B} \|_F }  \Big)_+ 
\end{align*}
where $k \ne \ell \in [p]$, $m \times m$ ${\bm B}$ has its $(u,v)$th element as $[{\bm B}]_{uv}=T_{st}([{\bm A}^{(k \ell)}]_{uv}, \alpha \lambda_{e,ij}/\rho^{(t)} )$, $i=(k-1)m+u$, $j=(\ell-1)m+v$.
\STATE Dual update ${\bm U}^{(t+1)} = {\bm U}^{(t)} + \left( \bm{\Omega}^{(t+1)} - {\bm V}^{(t+1)} \right)$. 
\STATE Check convergence. Set tolerances
\begin{align*}
   \tau_{pri} = & mp \, \tau_{abs} + \tau_{rel} \, \max ( \| {\bm \Omega}^{(t+1)} \|_F , \| {\bm V}^{(t+1)} \|_F ) \\
  \tau_{dual} = & mp \, \tau_{abs} + \tau_{rel} \,  \| {\bm U}^{(t+1)} \|_F / \rho^{(t)} \, .
\end{align*}
Define $d_p = \| {\bm \Omega}^{(t+1)} - {\bm V}^{(t+1)} \|_F$ and $d_d = \rho^{(t)} \| {\bm V}^{(t+1)} - {\bm V}^{(t)} \|_F$.
If $( d_p \le \tau_{pri}) \; \AND \; (d_d \le \tau_{dual})$, set converged = \TRUE .
\STATE Update penalty parameter $\rho$ $\,$ : 
\[
  \rho^{(t+1)} = \left\{ \begin{array}{ll} 2 \rho^{(t)} & \mbox{if  } d_p > \phi d_d \\  
	                                         \rho^{(t)} /2 & \mbox{if  } d_d > \phi d_p \\
																					 \rho^{(t)} & \mbox{otherwise} \, . \end{array} \right.
\]
We also need to set ${\bm U}^{(t+1)} = {\bm U}^{(t+1)}/2$ for $d_p > \phi d_d$ and ${\bm U}^{(t+1)} = 2 {\bm U}^{(t+1)}$ for $d_d > \phi d_p$.
\STATE $t \leftarrow t+1$
\ENDWHILE
\STATE For $k \ne \ell$, if $\|{\bm V}^{(k \ell)}\|_F > 0$, assign edge $\{ k, \ell\} \in \hat{\cal E}$, else $\{ k, \ell\} \not\in \hat{\cal E}$. Inverse covariance estimate $\hat{\bm \Omega} = {\bm V}$.
\end{algorithmic}
\end{algorithm}

Thus in LSP, with some initial guess $\bar{\bm{\Omega}}$, we replace 
\begin{align}
   \rho_\lambda ( | [{\bm \Omega}]_{ij} | )  & \rightarrow \lambda_{e,ij} :=
	 \frac{\lambda \epsilon }{| [\bar{\bm \Omega}]_{ij} |  +\epsilon} \, , \\
  \rho_\lambda ( \| {\bm \Omega}^{(k \ell)} \|_F )  & \rightarrow \lambda_{g,k \ell} :=
	 \frac{\lambda \epsilon }{\| \bar{\bm \Omega}^{(k \ell)} \|_F  +\epsilon} \, .
\end{align}
The solution $\hat{\bm{\Omega}}_{\rm lasso}$ to the convex sparse group-lasso-penalized objective function may be used as an initial guess with $\bar{\bm{\Omega}} = \hat{\bm{\Omega}}_{\rm lasso}$. Similarly, for SCAD, we have 
\begin{align} 
\lambda_{e,ij} = &
			 \left\{ \begin{array}{ll} \lambda , & \mbox{if  } | [\bar{\bm \Omega}]_{ij} | \le \lambda \\
				    \frac{ a \lambda - | [\bar{\bm \Omega}]_{ij} |}{ a-1} , 
									& \mbox{if  } \lambda < | [\bar{\bm \Omega}]_{ij} | \le a \lambda \\ 
						0 , & \mbox{if  } a \lambda < | [\bar{\bm \Omega}]_{ij} |  \end{array} \right.  \, , \\
  \lambda_{g,k \ell} = &
			 \left\{ \begin{array}{ll} \lambda , & \mbox{if  } \| \bar{\bm \Omega}^{(k \ell)} \|_F \le \lambda \\
				    \frac{ a \lambda - \| \bar{\bm \Omega}^{(k \ell)} \|_F}{ a-1} , 
									& \mbox{if  } \lambda < \| \bar{\bm \Omega}^{(k \ell)} \|_F \le a \lambda \\ 
						0 , & \mbox{if  } a \lambda < \| \bar{\bm \Omega}^{(k \ell)} \|_F  \end{array} \right.  \, .  
\end{align}
With LLA, the original objective function is transformed to its convex LLA approximation
\begin{align}  
   \tilde{\cal L}({\bm \Omega}) & =  {\cal L}({\bm \Omega}) + \alpha \tilde{P}_e({\bm \Omega}) 
	   + (1-\alpha) \tilde{P}_g({\bm \Omega}) , \label{admm} \\
	   \tilde{P}_e({\bm \Omega}) &  =  \sum_{i \ne j}^{mp} 
		   \lambda_{e,ij} \Big| [ \bm{\Omega}_k ]_{ij} \Big|  , \label{admm1} \\
		\tilde{P}_g({\bm \Omega}) &  = m  \, \sum_{ q \ne \ell}^p \; 
		    \lambda_{g,q \ell} \| {\bm \Omega}^{(q \ell )} \|_F \, . \label{admm2} 
\end{align}
For lasso, we have $\lambda_{e,ij} = \lambda$ $\forall i,j$ and $\lambda_{g,q \ell} = \lambda$ $\forall q, \ell$.
We follow an ADMM approach, as outlined in \cite{Tugnait21a}, for both lasso and LLA to LSP/SCAD. Consider the scaled augmented Lagrangian \cite{Boyd2010} for this problem after variable splitting, given by 
\begin{align}
 \bar{\cal L}_\rho &(\bm{\Omega} , {\bm V} , {\bm U}  ) =   
   {\cal L}( \bm{\Omega} )	+ \alpha \tilde{P}_e({\bm V}) 
	    \nonumber \\
		&  + (1-\alpha) \tilde{P}_g({\bm V}) + \frac{\rho}{2}  \| \bm{\Omega} - {\bm V} + {\bm U} \|^2_F \, , \label{admm0} 
\end{align}
where ${\bm V} \in \mathbb{R}^{(mp) \times (mp)}$ results from variable splitting, and in the penalties we use  ${\bm V}$ instead of ${\bm \Omega}$, adding the equality constraint ${\bm V}={\bm \Omega}$, $ {\bm U}$ is the dual variable,  and $\rho >0$ is the ``penalty parameter'' \cite{Boyd2010}. 

The main difference between \cite{Tugnait21a} and this paper is the fact that $P_g({\bm V})$ and $P_g({\bm \Omega})$ are penalized slightly differently in the two papers (the factor $m$ is missing from \cite{Tugnait21a}). For non-convex penalties (not considered in \cite{Tugnait21a}), we have an iterative solution: first solve with sparse group-lasso penalty, then use the LLA formulation and solve the resulting adaptive lasso type convex problem. In practice, just two iterations seem to be enough. A pseudo code for the ADMM algorithm used in this paper is given in Algorithm \ref{alg0} where we use the stopping (convergence) criterion following \cite[Sec.\ 3.3.1]{Boyd2010} and varying penalty parameter $\rho$ following \cite[Sec.\ 3.4.1]{Boyd2010}. See \cite{Tugnait21a} for further details; note that by construction, $\bm{\Omega}^{(t+1)}$ in step 4 of Algorithm \ref{alg0} is positive definite. Our ADMM-based optimization algorithm is as follows.
\begin{itemize}
\item[1.] Calculate sample covariance $\hat{\bm{\Sigma}}$ as in (\ref{moresm}). Initialize iteration $\tilde{m}=1$,  ${\bm \Omega}^{(0)} = (\mbox{diag}({\hat{\bm \Sigma}}))^{-1}$, $\bar{\bm{\Omega}}=  {\bm \Omega}^{(0)}$ and use $\bar{\bm{\Omega}}$ to compute $\lambda_{e,ij}$'s and $\lambda_{g,k \ell}$.
\item[2.] Execute Algorithm \ref{alg0} with initial guess $\bar{\bm{\Omega}}$. 
\item[3.] Quit if using sparse-group lasso, else set ${\bm \Omega}^{(\tilde{m})} = \hat{\bm \Omega}$ and $\bar{\bm{\Omega}} = {\bm \Omega}^{(\tilde{m})}$ to re-compute $\lambda_{e,ij}$'s and $\lambda_{g,q \ell}$'s via the LLA. Set $\tilde{m} \leftarrow \tilde{m}+1$.
\item[4.] Repeat steps 2 and 3 until convergence. The converged $\hat{\bm \Omega}$ is the final estimate of the inverse covariance. (For the numerical results shown in Sec.\ \ref{NE}, we terminated after two iterations of steps 2 and 3, similar to \cite{Zou2008, Lam2009}.)
\end{itemize}

\subsection{Convergence and Model Selection} \label{OI}
 In the LLA approach, each approximation yields a convex objective function, therefore, convergence to a global minimum of $\tilde{\cal L}({\bm \Omega})$ is guaranteed. Overall it is a majorization-minimization approach, hence, after repeated LLA's, one gets a local minimum of the original non-convex objective function. In practice, two iterations seem to be enough: first run  Algorithm \ref{alg0} for lasso, then using lasso-based LLA, run Algorithm \ref{alg0} once more.  

For model selection we follow the BIC information criterion as discussed in \cite{Tugnait21a}. Let $\hat{\bm{\Omega}}$ and $\hat{\bar{\cal E}}$ ($= \{ \{i,j\} \, : \, |V_{ij} | > 0, \; i \ne j\}$, ${\bm V}$ as in Algorithm \ref{alg0} after convergence) denote the estimated inverse covariance matrix and estimated enlarged edge-set, and let $| \hat{\bar{\cal E}} |$ denote the cardinality (\# of nonzero elements) of $\hat{\bar{\cal E}}$. Noting that $\hat{\bm{\Omega}}$ is symmetric with nonzero diagonal elements, the number of free nonzero elements of $\hat{\bm{\Omega}}$ equal $\frac{1}{2}| \hat{\bar{\cal E}} |$+$p m $. The BIC is then given by
\begin{equation}
    \mbox{BIC}(\lambda , \alpha) = {\rm tr} ( \hat{\bm{\Sigma}} \hat{\bm{\Omega}}  ) - \ln(|\hat{\bm{\Omega}}|) 
		     +  \frac{\ln (n)}{n} \, \Big( \frac{1}{2} | \hat{\bar{\cal E}} | \Big)
\end{equation}
based on the optimized negative log-likelihood $- \ln  f_{{\bm X}}({\bm X}) \propto  \frac{n}{2} \big( {\rm tr} ( \hat{\bm{\Sigma}}  \hat{\bm{\Omega}}  ) - \ln | \hat{\bm{\Omega}}| \big)$, where ${\bm X} = \{ {\bm x}(t) \}_{t=1}^{n}$ and $f_{{\bm X}}({\bm X})$ is the joint probability density function of ${\bm X}$.  The pair $(\lambda,\alpha)$ is selected to minimize BIC. Unlike \cite{Tugnait21a}, in this paper to simplify computations, we fix $\alpha = 0.05$ and select $\lambda$ to minimize $\mbox{BIC}(\lambda , 0.05)$ by searching over a grid of values for synthetic data. For real data, we search over both $\lambda$ and $\alpha$ as follows. Fix $\alpha = 0.05$ and select the best $\lambda$ by searching over a grid of values, and then with this optimized $\lambda$, select the best $\alpha$ by searching over a grid of values in $[0.01,0.3]$.
 
In our simulations we search over $\lambda \in [\lambda_{\ell} , \lambda_{u}]$,  where $\lambda_{\ell}$ and $\lambda_u$ are selected via a heuristic as in \cite{Tugnait21a}. Find the smallest $\lambda$, labeled $\lambda_{sm}$ for which we get a no-edge model; then we set $\lambda_{u}= \lambda_{sm}/2$ and $\lambda_{\ell} = \lambda_{u}/10$ for both synthetic and real datasets. For the numerical results presented in Sec.\ \ref{NE}, we picked $t_{\max} = 200$, $\bar{\rho} =2$, $\phi =10$, $\tau_{abs}=\tau_{rel} =10^{-4}$ in Algorithm \ref{alg0}. For the SCAD penalty $a=3.7$ (as in \cite{Lam2009}) and for the log-sum penalty $\epsilon = 0.0001$.

\section{Theoretical Analysis} \label{TA}
Here we analyze the properties of $\hat{\bm{\Omega}} = \arg\min_{\bm{\Omega}  \succ {\bm 0}}  \bar{\cal L}(\bm{\Omega})$. Since the SCAD and log-sum penalties are non-convex, the objective function is  non-convex and in general, any optimization of the objective function will yield only a stationary point or a local minimum.
We now allow $p$ and $\lambda$ to be functions of sample size $n$, denoted as $p_n$ and $\lambda_n$, respectively. 
Recall that we have the original multi-attribute graph ${\cal G} = (V, {\cal E})$ with $|V|=p_n$ and the corresponding enlarged graph $\bar{\cal G} = (\bar{V}, \bar{\cal E})$ with $|\bar{V}|=mp_n$. 

\subsection{Analysis without Irrepresentability Conditions} \label{TA1}
We assume the following regarding ${\cal G}$.
\begin{itemize}
\item[(A1)] Denote the true edge set of the graph by ${\cal E}^\ast$, implying that ${\cal E}^\ast = \{ \{j,k\} ~:~ \|({\bm \Omega}^\ast)^{(jk)}\|_F > 0, ~j\ne k \}$ where ${\bm \Omega}^\ast$ denotes the true precision matrix of ${\bm x}(t)$.  Assume that card$({\cal E}^\ast) =|{\cal E}^\ast| \le s_{n}^\ast$.

\item[(A2)] The minimum and maximum eigenvalues of $(mp_n) \times (mp_n)$ true covariance $\bm{\Sigma}^\ast  \succ {\bm 0}$  satisfy 
\begin{align*}
     0 < \beta_{\min} & \le \phi_{\min}(\bm{\Sigma}^\ast) \le 
		     \phi_{\max}(\bm{\Sigma}^\ast) \le \beta_{\max} < \infty \, .
\end{align*}
Here $\beta_{\min}$ and $\beta_{\max}$ are not functions of $n$ (or $p_n$).
\end{itemize}

Let $\hat{\bm{\Omega}} = \arg\min_{\bm{\Omega}  \succ {\bm 0}}  \bar{\cal L}(\bm{\Omega})$.
Theorem 1  establishes local consistency of $\hat{\bm{\Omega}}$ (the proof is in Appendix \ref{append1}). \\
{\it Theorem 1 (Local Consistency)}. For $\tau > 2$, let
\begin{align}  
    C_0 = & 40 \, \max_{k \in [mp]} ( [{\bm \Sigma}^\ast]_{kk} ) 
	    \sqrt{ N_1 / \ln ( p_n )  }  \, ,  \label{naeq58} \\ 
    R = & 8(1+m)C_0 / \beta_{\min}^2 \, , \label{neq15ab0} \\
    r_n = & \sqrt{ (m p_n+ m^2 s_{n}^\ast) \ln ( p_n )/n} = o(1)\, , \label{neq15ab1} \\
		N_1 = &  2 \ln (4 m^2 p_n^\tau )  \, ,  \label{neq15ab2} \\
		N_2 = & \arg \min \big\{ n \, : \, r_n \le 
		     0.1/ ( R \beta_{\min} ) \big\} \, , \label{neq15ab3} \\
    N_3 = &  \arg \min \Big\{ n \, : \, r_n \le \frac{\epsilon}{R} \Big\} \, ,  \label{neq15ab4}  \\
		N_4 = &  \arg \min \Big\{n \, : \, \lambda_n \le 
		       \frac{\min_{(i,j): \, [{\bm \Omega}^\ast]_{ij} \ne 0} 
					 |[{\bm \Omega}^\ast]_{ij}| }{a+1} \Big\}   \, ,  \label{neq15ab41}  \\
		\lambda_{n \ell} = & 2 C_0  \sqrt{\ln (p_n )/n} \, ,  \label{neq15ab5} \\
		\lambda_{n u1} = &  C_0 (m+1) r_n /(m \sqrt{s_{n}^\ast})  \, , \label{neq15ab6} \\
		\lambda_{n u2} = & \min \left( R r_n , \lambda_{n u1} \right) \, .  \label{neq15ab7} 
\end{align}
Under assumptions (A1)-(A2), there exists a local minimizer $\hat{\bm{\Omega}}$ of $\bar{\cal L}(\bm{\Omega})$ satisfying
\begin{equation}  \label{neq15}
  \| \hat{\bm{\Omega}} - \bm{\Omega}^\ast \|_F  \le R  r_n
\end{equation}
with probability greater than $1-1/p_n^{\tau-2}$ if
\begin{itemize}
\item[(i)] for the lasso penalty  $n >  \max \{ N_1, N_2 \}$ and $\lambda_n$ satisfies $\lambda_{n \ell} \le \lambda_n \le \lambda_{nu1}$,
\item[(ii)] for the SCAD penalty  $n >  \max \{ N_1, N_2, N_4 \}$ and $\lambda_n$ satisfies $\lambda_n = \lambda_{nu2}$,
\item[(iii)] for the log-sum penalty  $n >  \max \{ N_1, N_2, N_3 \}$ and $\lambda_n$ satisfies $\lambda_{n \ell} \le \lambda_n \le \lambda_{nu1}$.
\end{itemize}
For the lasso penalty, $\hat{\bm{\Omega}}$ is a global minimizer whereas for the other two penalties, it is a local minimizer. $\quad \bullet$ \\
{{\it Remark 1}. {\it Convergence Rate}}. In terms of the rate of convergence,  
  $\| \hat{\bm{\Omega}} - \bm{\Omega}^\ast \|_F 
	        = {\cal O}_P \left( r_n \right) = {\cal O}_P \left( r_n/m \right)$ for fixed $m$. 
Therefore, for $\| \hat{\bm{\Omega}} - \bm{\Omega}^\ast \|_F \rightarrow 0$ as $n \rightarrow \infty$, we must have $(p_n m^{-1}+ s_{n}^\ast) \ln ( p_n )/n \rightarrow 0$. Notice that $m p_n+ m^2 s_{n}^\ast$ is the maximum number of nonzero elements in $\bm{\Omega}^\ast$.  $\;\; \Box$

We follow the proof technique of \cite[Lemma 6]{Loh2017} in establishing  Lemma 1 (the proof is in Appendix \ref{append2}). \\
{\it Lemma 1 (Local Convexity)}. The optimization problem
\begin{align}  
    \hat{\bm{\Omega}} = & \arg\min_{ \bm{\Omega} \in {\cal B} }  
		\bar{\cal L}(\bm{\Omega}) \, , \label{neq4000} \\
		{\cal B} = & \{ \bm{\Omega} \,:\, \bm{\Omega} \succ {\bm 0} , \; 
		   \| \bm{\Omega} \| \le 0.99 \; \bar{\mu} \} \, , \\
		\bar{\mu} = & \left\{ \begin{array}{ll}
		   \infty & : \;\; \mbox{lasso} \\
			  \sqrt{(a-1)/m} & : \;\; \mbox{SCAD} \\
				\sqrt{ \epsilon / (m \lambda_n) } & : \;\; \mbox{log-sum}, \end{array} \right. \label{neq4010} 
\end{align}
consists of a strictly convex objective function over a convex constraint set for all three penalties where $\lambda_n$ is as in Theorem 1. $\quad \bullet$

Lemma 1 and Theorem 1 lead to Theorem 2, as proved in Appendix \ref{append2}. \\
{\it Theorem 2}. Assume the conditions of Theorem 1. If $R r_n + 1/\beta_{\min} \le 0.99 \; \bar{\mu}$, then $\hat{\bm{\Omega}}$ as defined in Lemma 1 is a unique minimum, satisfying all results of Theorem 1.  $\quad \bullet$

{\it Remark 2}. We see from Theorem 1 that as $n \rightarrow \infty$, $\lambda_n \rightarrow 0$ (since $r_n = o(1)$), therefore, we eventually have ``global'' convexity for log-sum penalty by (\ref{neq4010}) for any $\bm{\Omega}^\ast$. But such is not the case for SCAD where one may need $a$ to become large in which case it would behave more like lasso. $\quad \Box$

We now turn to graph recovery. Define
\begin{align}  
    \hat{\cal E} = & \left\{ \{q,\ell\} \, : \, \|\hat{\bm \Omega}^{(q \ell)} \|_F > \theta_n > 0,
		    q \ne \ell \right\} \, , \label{neq4020} \\
		{\cal E}^\ast = &   \left\{ \{q,\ell\} \, : \, \|({\bm \Omega}^\ast)^{(q \ell)} \|_F >  0,
		    q \ne \ell \right\} \, , \label{neq4022} \\
       \bar{\sigma}_n = &   R r_n   \, , \label{neq4022} \\
			\nu  = &  \min_{\{q,\ell\} \in {\cal E}^\ast} \,  \|({\bm \Omega}^\ast)^{(q \ell)} \|_F  \, , \label{neq4024} \\
			N_4 = & \arg \min \Big\{ n \, : \, \bar{\sigma}_n \le 0.4 \nu \Big\} \, , \label{neq4025}
\end{align}
where $R$ and $r_n$ are as in (\ref{neq15ab0}) and (\ref{neq15ab1}), respectively. We follow the proof technique of \cite[Theorem 10]{Zhao2022} in establishing  Theorem 3 whose proof is in Appendix \ref{append2}.  \\
{\it Theorem 3}. For $\theta_n = 0.5 \nu$ and $n \ge N_4$, $\hat{\cal E} = {\cal E}^\ast$ with probability$> \, 1-1/p_n^{\tau-2}$ under the conditions of Theorem 1.  $\quad \bullet$

{\it Remark 3}.  In practice we do not know the value of $\nu$, hence cannot calculate $\theta_n$ needed in (\ref{neq4020}). For the numerical results presented in Sec.\ \ref{NE}, we used $\theta_n =0$. Using some irrepresentability conditions (not needed in Theorem 1) and the primal-dual witness method, in Theorem 4(iv) of Sec.\ \ref{TA2} we establish a result similar to Theorem 3 but with $\theta_n =0$. That is, additional sufficient conditions on the system model lead to sharper results in Sec.\ \ref{TA2}.
$\quad \Box$

\subsection{Analysis With Irrepresentability Conditions} \label{TA2}
Here we impose additional conditions and obtain sharper results. Now we follow the primal-dual witness technique of \cite{Ravikumar2011}, originally used in the context of element-wise lasso penalty for single-attribute graphs. The technique of \cite{Ravikumar2011} was extended to group-lasso penalty for multi-attribute graphs in  \cite{Kolar2014}. In this paper we apply the primal-dual witness technique in a sparse-group non-convex penalty setting. Some prior results in an element-wise non-convex penalty setting are in \cite{Loh2015, Loh2017}.

Denote the true extended graph edgeset $\bar{\cal E}$ by $\bar{\cal E}^\ast$. Define 
\begin{align}
 S = & {\cal E}^\ast \cup \big\{ \{k, \ell\} \, : \, k=\ell \in [p_n] \big\} \subseteq [p_n] \times [p_n]  \, ,
    \label{meq100}  \\
 \bar{S} = & \bar{\cal E}^\ast \cup \big\{ \{i,j\} \, : \, i = j \in [mp_n] \big\} \subseteq [mp_n] \times [mp_n]  \, ,
    \label{meq102}  \\
 \Gamma^\ast = & ({\bm \Omega}^\ast)^{-1} \boxtimes ({\bm \Omega}^\ast)^{-1}\, , \label{meq110}  \\
 \hat{\Gamma} = & \hat{\bm \Omega}^{-1} \boxtimes \hat{\bm \Omega}^{-1} \, , \label{meq120}   \\
\kappa_{\Gamma^\ast} = & \| \bm{\mathcal C}((\Gamma_{S,S}^\ast)^{-1}) \|_{1, \infty} \, \label{meq130}   \\
{\bar \kappa}_{\Gamma^\ast} = & \| (\Gamma_{S,S}^\ast)^{-1} \|_{1, \infty} \, \label{meq140}   \\
\kappa_{\Sigma^\ast} = & \| \bm{\mathcal C}(\Sigma^\ast) \|_{1, \infty} \, \label{meq150}   \\
{\bar \kappa}_{\Sigma^\ast} = & \| \Sigma^\ast \|_{1, \infty} \, \label{meq160}   \\
 d_n = & \mbox{maximum degree of } S \, , \label{meq170}   \\
 \bar{d}_n = & \mbox{maximum degree of } \bar{S} . \label{meq180}  
\end{align}
In (\ref{meq170}), $d_n$ is the maximum number of non-zero elements per row of $\bm{\mathcal C}(\Omega^\ast)$, and similarly $\bar{d}_n$ in (\ref{meq180}) is the maximum number of non-zero elements per row of $\Omega^\ast$.  As discussed in Sec.\ \ref{SM}, with the true graph ${\cal G}^\ast = (V, {\cal E}^\ast)$, $V = [p_n]$, we associate an enlarged graph $\bar{\cal G}^\ast = (\bar{V}, \bar{\cal E}^\ast)$, $\bar{V} = [mp_n]$, such that corresponding to an edge $f = \{ k, \ell\} \in  {\cal E}^\ast$, there are $m^2$ edges $\{i,j \} \in \bar{\cal E}^\ast$ specified by $i=(k-1)m+q$, $j=(\ell-1)m+r$, $q,r \in [m]$. To keep notation light, for an edge $f = \{ k, \ell\} \in  {\cal E}^\ast$, we use $e_f$ to denote an edge $\{i,j \} \in \bar{\cal E}^\ast$ that corresponds to one of $m^2$ edges of $\bar{\cal E}^\ast$ associated with edge $f$. Similar notation will be used for edges in $({\cal E}^\ast)^c$ and  $(\bar{\cal E}^\ast)^c$, and  edges in $S$ and $S^c$. Using this notation, assume that for some $\gamma \in (0,1]$, the following {\it irrepresentability} conditions hold:
\begin{align}
  \max_{f \in S^c} & \| \bm{\mathcal C}({\bm \Gamma}_{f,S}^\ast ({\bm \Gamma}_{S,S}^\ast)^{-1}) \|_{1} \, \le \, 1- \gamma 
	   \, , \label{meq200}   \\
	\max_{e_f \in f \in S^c} & \| {\bm \Gamma}_{e_f,S}^\ast ({\bm \Gamma}_{S,S}^\ast)^{-1} \|_{1} \, \le \, 1- \gamma \, .
	  \label{meq210}  
\end{align}

\begin{table*}
\vspace*{-0.1in}
\caption{\small{\it $F_1$ scores, Hamming distances, normalized Frobenius norm of estimation error ($\| \hat{\bm \Omega} - {\bm \Omega}^\ast \|_F / \| {\bm \Omega}^\ast \|_F$), and timing, for the synthetic data examples ($p=100$, $m=4$), averaged over 100 runs (standard deviation $\sigma$ in parentheses).}} \label{table1} 
\vspace*{-0.15in}
\begin{center}
\begin{tabular}{c|ccc|ccc}   \hline\hline
 $n$ &  200 &  400  & 800  &  200 &  400  & 800  \\  \hline\hline
\multicolumn{7}{c}{ $\lambda$'s picked to maximize $F_1$ score } \\
\multicolumn{4}{c|}{ER graph: $F_1$ score ($\sigma$) } 
 & \multicolumn{3}{c}{BA graph: $F_1$ score ($\sigma$)}\\ \hline
Lasso  &  0.742 (0.065)  &  0.916 (0.032)  & 0.983 (0.011)   
  & 0.573 (0.058)  &  0.784 (0.067)  & 0.918 (0.048) \\
Log-sum   &  0.804 (0.046)  &  0.964 (0.008)  & 0.998 (0.002)   
  & 0.647 (0.053)  &  0.886 (0.044)  & 0.983 (0.017) \\ 
	SCAD   &  0.752 (0.072)  &  0.931 (0.017)  & 0.988 (0.007)   
  & 0.590 (0.062)  &  0.800 (0.075)  & 0.933 (0.053) \\ \hline\hline
	\multicolumn{4}{c|}{ER graph: Hamming distance ($\sigma$) } 
 & \multicolumn{3}{c}{BA graph: Hamming distance ($\sigma$)}\\ \hline
Lasso  &  113.4 (18.24)  &  39.60 (13.78)  & 08.49 (05.50)   
  & 181.4 (68.10)  &  80.66 (18.54)  & 31.38 (14.98) \\
Log-sum   &  86.91 (17.52)  &  18.01 (04.26)  & 0.880 (0.898)   
  & 128.3 (10.15)  &  41.32 (13.59)  & 06.45 (06.14) \\ 
	SCAD   &  105.7 (21.39)  &  34.63 (08.74)  & 06.16 (03.48)   
  & 161.6 (46.10)  &  71.55 (19.23)  & 24.89 (16.26) \\ \hline\hline
		\multicolumn{4}{c|}{ER graph: Est.\ error ($\sigma$) } 
 & \multicolumn{3}{c}{BA graph: Est.\ error ($\sigma$) }\\ \hline
Lasso  &  0.335 (0.008)  &  0.303 (0.010)  & 0.266 (0.010)   
  & 0.268 (0.005)  &  0.241 (0.003)  & 0.212 (0.005) \\
Log-sum   &  0.307 (0.008)  &  0.227 (0.008)  & 0.170 (0.007)   
  & 0.265 (0.004)  &  0.214 (0.004)  & 0.164 (0.006) \\ 
	SCAD   &  0.313 (0.008)  &  0.222 (0.007)  & 0.149 (0.005)   
  & 0.304 (0.028)  &  0.217 (0.004)  & 0.152 (0.010) \\ \hline\hline
			\multicolumn{4}{c|}{ER graph: Timing (s) ($\sigma$) } 
 & \multicolumn{3}{c}{BA graph: Timing (s) ($\sigma$)}\\ \hline
Lasso  &  1.897 (0.140) & 1.895 (0.251) & 1.891 (0.059)   
  & 2.038 (0.358) & 1.932 (0.315) & 1.958 (0.277) \\
Log-sum   &  7.604 (0.819)  &  6.251 (0.185)  & 5.765 (0.228)   
  & 6.902 (0.236) & 6.431 (0.263) & 5.857 (0.341) \\ 
	SCAD   &  5.158 (0.455) & 5.291 (0.243) & 5.027 (0.237)   
  & 6,574 (0.808) & 5.473 (0.437) & 5.133 (0.302) \\ \hline\hline
	\multicolumn{7}{c}{ $\lambda$'s picked to minimize BIC } \\
\multicolumn{4}{c|}{ER graph: $F_1$ score ($\sigma$) } 
 & \multicolumn{3}{c}{BA graph: $F_1$ score ($\sigma$)}\\ \hline
Lasso  &  0.329 (0.147) & 0.813 (0.085) & 0.965 (0.039)   
  & 0.414 (0.149) & 0.622 (0.158) & 0.901 (0.068) \\
Log-sum   &  0.766 (0.083) & 0.942 (0.021) & 0.996 (0.004)   
  & 0.582 (0.120)  &  0.859 (0.076)  & 0.936 (0.099) \\ \hline\hline
	\multicolumn{4}{c|}{ER graph: Hamming distance ($\sigma$) } 
 & \multicolumn{3}{c}{BA graph: Hamming distance ($\sigma$)}\\ \hline
Lasso  &  1239. (538.2) & 125.6 (167.3) & 019.1 (023.2)   
  & 739.1 (781.3) & 255.3 (204.9) & 36.15 (19.49)  \\
Log-sum   &  136.3 (70.81) & 27.54 (10.01) & 01.88 (01.64)   
  & 240.1 (185.7) & 48.29 (20.05) & 20.91 (30.22) \\  \hline\hline
\end{tabular} 
\vspace*{-0.1in}
\end{center}
\end{table*} 

With $C_0$ and $N_1$ as defined in (\ref{naeq58}) and (\ref{neq15ab2}), respectively, define $\tilde{C}_0 = m C_0$ and 
\begin{align}
 N_5 = & 36 d_n^2 \kappa_{\Gamma^\ast}^2 \Big(1+\frac{4}{\gamma} \Big)^2 \tilde{C}_0 \ln (p_n)
  \, \max \big\{ \kappa_{\Sigma^\ast}^2 , \kappa_{\Gamma^\ast}^2  \kappa_{\Sigma^\ast}^6 \big\} \, , \label{meq300}  \\
 N_6 = & 36 d_n^2  \Big(1+\frac{4}{\gamma} \Big)^4 \tilde{C}_0^2 \ln (p_n)
  \, \kappa_{\Gamma^\ast}^4  \kappa_{\Sigma^\ast}^6  \, ,  \label{meq310}  \\
 N_7 = & 36 {\bar d}_n^2  \Big(1+\frac{4}{\gamma} \Big)^4 m^2 \tilde{C}_0^2 \ln (p_n)
  \, \kappa_{\Gamma^\ast}^4  \kappa_{\Sigma^\ast}^6  \, . \label{meq320}     
\end{align}
Let $\partial \bar{\cal L}({\bm \Omega})$ denote the sub-differential of $\bar{\cal L}({\bm \Omega})$ at ${\bm \Omega}$. Suppose that $\hat{\bm \Omega}$ is a solution to
\begin{align}
  {\bm 0} \, \in & \, \partial \bar{\cal L}({\bm \Omega}) \, , \label{meq220}
\end{align}
which is a first-order necessary condition for a stationary point of $\bar{\cal L}({\bm \Omega})$.  
Theorem 4 addresses some properties of this $\hat{\bm \Omega}$. \\
{\it Theorem 4}. For the system model of Sec.\ \ref{SM}, under the irrepresentability conditions (\ref{meq200})-(\ref{meq210}) for some $\gamma \in (0,1]$, if
\begin{align}  
   \lambda_n = & \frac{4}{\gamma} \, C_0 \, \sqrt{\frac{\ln (p_n)}{n}} \, ,  \label{eqn350} 
\end{align}
then for $n > \max (N_1, N_5, N_6, N_7)$ and for any $\tau >2$, there exists a stationary point $\hat{\bm \Omega}$ of $\bar{\cal L}({\bm \Omega})$ satisfying  with probability $> 1- 1/p_n^{\tau -2}$, 
\begin{itemize}
\item[(i)] $\| \bm{\mathcal C}(\hat{\bm \Omega} - {\bm \Omega}^\ast) \|_\infty \le 
  2 \kappa_{\Gamma^\ast} (1+\frac{4}{\gamma}) m C_0 \sqrt{\frac{\ln (p_n)}{n}} $,
\item[(ii)] $\hat{\bm \Omega}_{S^c} = {\bm 0}$.
\item[(iii)] $\| \bm{\mathcal C}(\hat{\bm \Omega} - {\bm \Omega}^\ast) \|_F \le \sqrt{s_n^\ast + p_n} \, \| \bm{\mathcal C}(\hat{\bm \Omega} - {\bm \Omega}^\ast) \|_\infty$ .
\item[(iv)] Additionally, if $\min_{(k,\ell) \in S} \| ({\bm \Omega}^\ast)^{(k \ell)} \|_F \ge  \newline 4 \kappa_{\Gamma^\ast} (1+\frac{4}{\gamma}) m C_0 \sqrt{\frac{\ln (p_n)}{n}}$, then $P(\hat{\cal E} = {\cal E}^\ast) > 1- 1/p_n^{\tau -2}$  where $\hat{\cal E} =  \left\{ \{q,\ell\} \, : \, \|\hat{\bm \Omega}^{(q \ell)} \|_F >  0,
		    q \ne \ell \right\}$. $\quad \bullet$
\end{itemize}
The proof of Theorem 4 is given in Appendix \ref{append3}.

Lemma 1 and Theorem 4 lead to Theorem 5, as proved in Appendix \ref{append3}. First define
\begin{align}
  \tilde{R} & = 2 \kappa_{\Gamma^\ast} \Big(1+\frac{4}{\gamma} \Big) m C_0 \, , \\
	\tilde{r}_n & = \sqrt{(s_n^\ast + p_n) \frac{\ln (p_n) }{ n }} \, .
\end{align}
{\it Theorem 5}. Assume the conditions of Theorem 4, and as in assumption (A2), suppose that $\beta_{\min}  \le \phi_{\min}(\bm{\Sigma}^\ast)$. If $\tilde{R}  \tilde{r}_n + 1/\beta_{\min} \le 0.99 \; \bar{\mu}$, then $\hat{\bm{\Omega}}$ as defined in Lemma 1 is a unique minimum, satisfying all results of Theorem 4. $\quad \bullet$

{{\it Remark 4}. In terms of the rate of convergence,  
  $\| \bm{\mathcal C}(\hat{\bm \Omega} - {\bm \Omega}^\ast) \|_F 
	        = {\cal O}_P \left( \tilde{r}_n  \right)$. 
Therefore, for $\| \bm{\mathcal C}(\hat{\bm \Omega} - {\bm \Omega}^\ast) \|_F \rightarrow 0$ as $n \rightarrow \infty$, we must have $(p_n + s_{n}^\ast) \ln ( p_n )/n \rightarrow 0$. This is similar to the results of Theorem 1 (see Remark 1). But unlike Theorem 1, by Theorem 4(ii), we have the oracle result: $\hat{\bm \Omega}_{S^c} = {\bm 0}$ just as ${\bm \Omega}_{S^c}^\ast = {\bm 0}$, i.e., all absent edges in the true graph are absent in the estimated graph with high probability. Such a result does not exist for Theorem 1. Also, the comments made in Remark 2 apply here as well. Finally, Theorem 4 does not need a minimum amplitude condition like (\ref{neq15ab41}) in Theorem 1 for SCAD.  $\;\; \Box$

\section{Numerical Examples} \label{NE}
We now present numerical results for synthetic as well as real data to illustrate the proposed non-convex penalty approaches. In the synthetic data examples the ground truth is known which allows for an assessment of the efficacy of various approaches. In the real data example our goal is visualization and exploration of the conditional dependency structure underlying the data since the ground truth is unknown.

\begin{figure*}[ht]
\begin{subfigure}[t]{.5\textwidth}
  \centering
  \includegraphics[width=.8\linewidth]{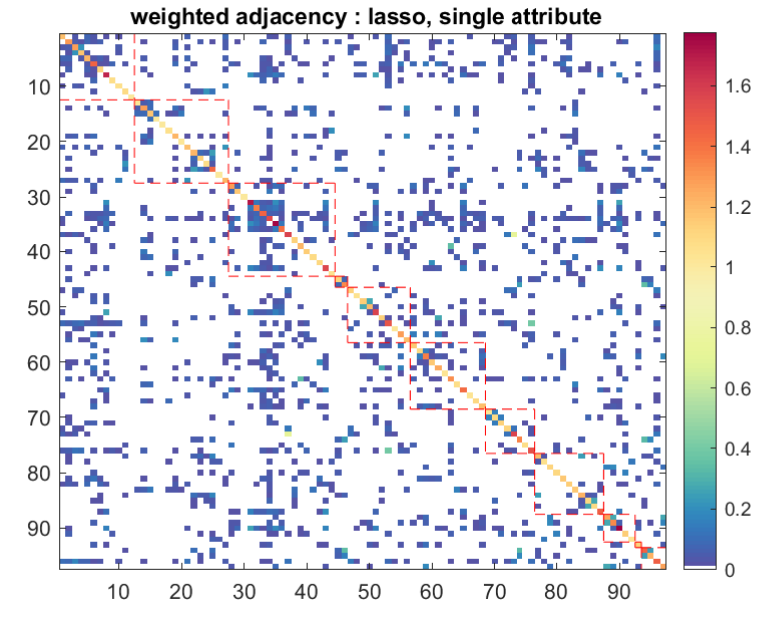}  
  \caption{Estimated $|\Omega_{k \ell}|$, lasso; 1021 edges}
  \label{fig1a}
\end{subfigure}%
\begin{subfigure}[t]{.5\textwidth}
  \centering
  \includegraphics[width=.8\linewidth]{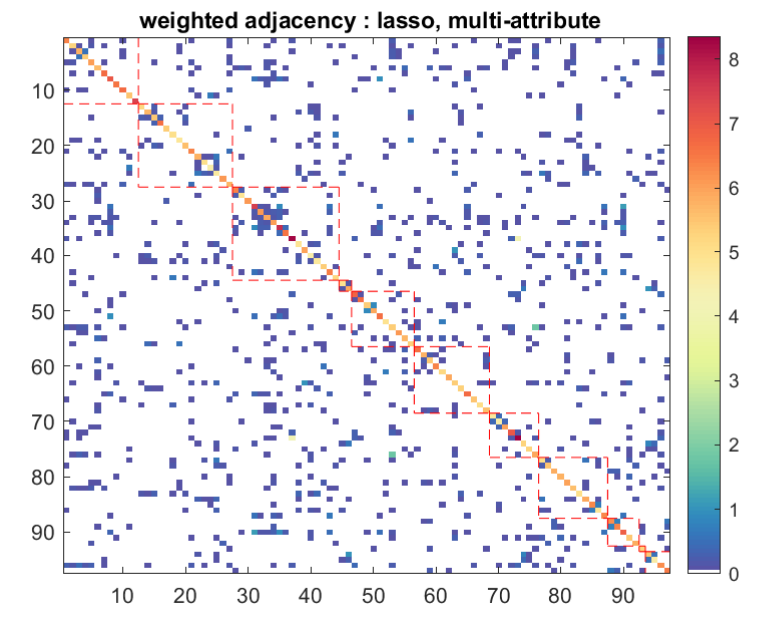}  
  \caption{Estimated $\| \hat{\bm \Omega}^{(k \ell)} \|_F$, lasso; 1517 edges.}
  \label{fig1b}
\end{subfigure}
\begin{subfigure}[t]{.5\textwidth}
  \centering
  \includegraphics[width=.8\linewidth]{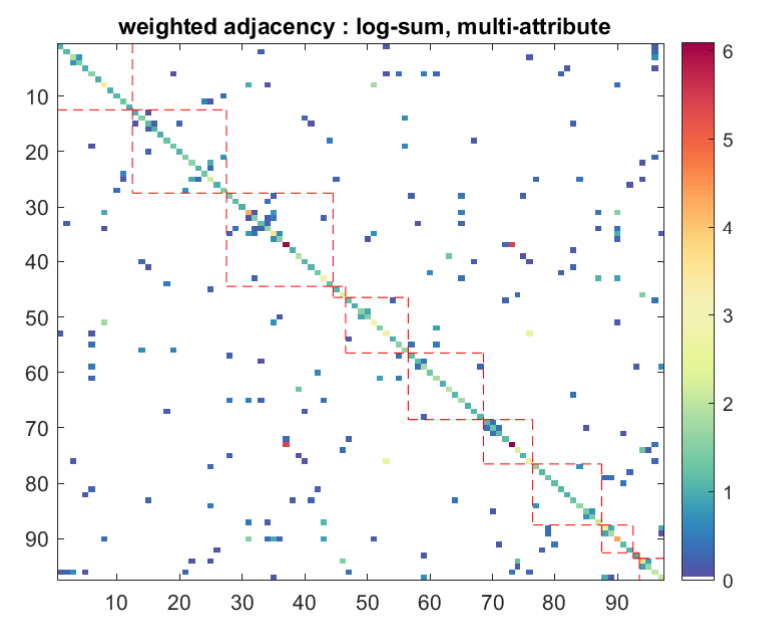}  
  \caption{Estimated $|\Omega_{k \ell}|$, log-sum; 128 edges}
  \label{fig1c}
\end{subfigure}%
\begin{subfigure}[t]{.5\textwidth}
  \centering
  \includegraphics[width=.8\linewidth]{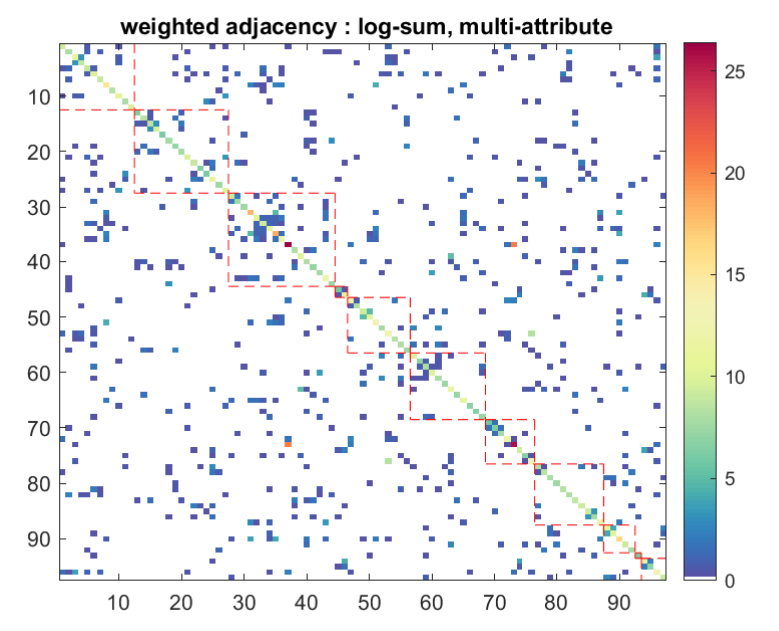}  
  \caption{Estimated $\| \hat{\bm \Omega}^{(k \ell)} \|_F$, log-sum; 684 edges.}
  \label{fig1d}
\end{subfigure}
\caption{Estimated precision matrices based edge weights for financial time series using proposed log-sum penalty with BIC for $\lambda$ selection. (a) Lasso, single-attribute case ($m=1$, $p=97$) with $|[\hat{\bm \Omega}]_{k \ell}|$, $k,\ell \in [97]$ as edge weight. (b) Lasso, multi-attribute case ($m=4$, $p=97$) with $\| \hat{\bm \Omega}^{(k \ell)} \|_F$, $k,\ell \in [97]$ as edge weight. (c) Log-sum, single-attribute case ($m=1$, $p=97$) with $|[\hat{\bm \Omega}]_{k \ell}|$, $k,\ell \in [97]$ as edge weight. (d) Log-sum, multi-attribute case ($m=4$, $p=97$) with $\| \hat{\bm \Omega}^{(k \ell)} \|_F$, $k,\ell \in [97]$ as edge weight. In all figures, the red squares (in dashed lines) show the 11 sectors -- they are not part of the edge weights.}
\label{fig1}
\end{figure*}

\subsection{Synthetic Data: Erd\"{o}s-R\`{e}nyi and Barab\'{a}si-Albert Graphs} \label{NEsyn}
We consider two types of graphs: Erd\"{o}s-R\`{e}nyi (ER) graph and Barab\'{a}si-Albert (BA) graph \cite{Barabasi1999, Lu2014}.  In the ER graph, $p=100$ nodes are connected to each other with probability $p_{er} =0.05$ and there are $m=4$ attributes per node whereas in the BA graph, we used $p=100$ and mean degree of 2 to generate a BA graph using the procedure given in \cite{Lu2014}. In the upper triangular $\bm{\Omega}$, we set $[\bm{\Omega}^{(jk)}]_{st} = 0.5^{|s-t|}$ for $j=k \in [p]$, $s,t \in [m]$. For $j \ne k$, if the two nodes are not connected in the graph (ER or BA), we have  $\bm{\Omega}^{(jk)} = {\bm 0}$, and if nodes $j$ and $k$ are connected, then $[\bm{\Omega}^{(jk)}]_{st}$ is uniformly distributed over $[-0.4,-0.1] \cup [0.1,0.4]$ for $s \ne t$, otherwise it is zero. Then add lower triangular elements to make $\bm{\Omega}$ a symmetric matrix. Finally add $\delta {\bm I}$ to  $\bm{\Omega}$ and pick $\delta$ so that the minimum eigenvalue of $\bm{\Omega}$ is 0.5; this is similar to the simulation example 3 in \cite[Sec.\ 5.1]{Kolar2014}.  With $\bm{\Phi} \bm{\Phi}^\top =(\bm{\Omega}+\delta {\bm I})^{-1}$, we generate ${\bm x} = \bm{\Phi} {\bm w}$ with ${\bm w} \in \mathbb{R}^{mp}$ as zero-mean Gaussian, with identity covariance. We generate $n$ i.i.d.\ observations for ${\bm x}$, with $m=4$, $p =100$, $n \in \{200, 400, 800\}$.

Simulation results based on 100 runs are shown in Table \ref{table1} for ER and BA graphs where the performance measures are $F_1$-score and Hamming distance (between estimated and true edgesets $\hat{\cal E}$ and ${\cal E}^\ast$) for efficacy in edge detection, normalized estimation error $\| \hat{\bm \Omega} - {\bm \Omega}^\ast \|_F / \| {\bm \Omega}^\ast \|_F$ and execution time (based on tic-toc functions in MATLAB). All simulations were run on a Window 10 Pro operating system with processor Intel(R) Core(TM) i7-10700 CPU @2.90 GHz with 32 GB RAM, using MATLAB R2023a. We used the ADMM algorithm (with LLA for non-convex penalties) given in Algorithm \ref{alg0} with $\alpha = 0.05$ for all three regularizations: lasso, log-sum and SCAD. It is seen that log-sum penalty outperforms lasso and SCAD with $F_1$ score or Hamming distance as the performance metric. For $n = 800$, SCAD yields smaller estimation errors in estimating ${\bm \Omega}$ but its performance in terms of $F_1$ score and Hamming distance metrics is, in general, poor. In practice we do not know the ground truth, hence cannot pick $\lambda$ to maximize the $F_1$ score. In Table \ref{table1} we also show results for lasso and log-sum penalties when $\lambda$ is picked based the BIC information criterion as discussed in Sec.\ \ref{OI} with $\alpha = 0.05$. Here again the log-sum penalty outperforms lasso.

\subsection{Real data: Financial Time Series} \label{NEreal}
We consider daily high, low and close-of-the-day share prices and daily trade volume ($m=4$) of 97 ($p=97$) stocks in the S\&P 100 index from April 1, 2015 through April 1, 2020, yielding 1259 samples. This data was gathered from the Yahoo Finance website. Let $[{\bm z}_i(t)]_\ell$ denote the $\ell$th feature ($\ell \in [m]$) of the $i$th stock on day $t$. Then we consider (as is conventional in such studies) $[{\bm x}_i(t)]_\ell = \ln ([{\bm z}_i(t)]_\ell/[{\bm z}_i(t)]_\ell)$ as the time series to analyze, yielding $n=1258$, $m=4$ and $p=97$. These 97 stocks are classified into 11 sectors (according to the Global Industry Classification Standard (GICS)) and we order the nodes to group them as information technology (nodes 1-12), health care (13-27), financials (28-44), real estate (45-46), consumer discretionary (47-56), industrials (57-68), communication services (69-76), consumer staples (77-87), energy (88-92), materials (93), utilities (94-97). For each $i$ and $\ell$, $[{\bm x}_i(t)]_\ell$ was centered and normalized to unit variance. We applied the BIC criterion for $\lambda$ selection using log-sum penalty as well as lasso, for two cases: $m=4$, and $m=1$ where only the close-of-the-day share prices were used. To apply BIC, with $\alpha =0.05$, we selected the best $\lambda$ as detailed in Sec.\ \ref{OI}, and then using this optimized $\lambda$, we selected the best $\alpha$ by searching over a grid of values in $[0.01,0.3]$. The resulting estimated precision matrices based edge weights are shown in Fig.\ 1. In the multi-attribute case ($m=4$, Figs.\ \ref{fig1b} and \ref{fig1d}), we get 1517 and 684 edges for lasso and log-sum regularizations, respectively, whereas in the single-attribute case ($m=1$, Figs.\ \ref{fig1a} and \ref{fig1c}) we get 1021 and 128 edges for lasso and log-sum regularizations, respectively. The multi-attribute graph exploits many more relevant features compared to the single-attribute graph which cannot use more than one feature. The log-sum penalty yields sparser graphs for both single-attribute and multi-attribute cases, compared to the lasso penalty.

\section{Conclusions}
We investigated multi-attribute graph learning using a penalized log-likelihood objective function where both convex (sparse-group lasso) and non-convex (sparse-group log-sum and SCAD) regularization functions were considered. An ADMM approach coupled with a local linear approximation to non-convex penalties was presented for optimization of the objective function. We established  sufficient conditions in a high-dimensional setting for consistency, local convexity when using non-convex penalties, and graph recovery. Two alternative sets of sufficient conditions were investigated, with and without some irrepresentability conditions. With irrepresentability conditions we could establish sharper results such as the oracle property (Theorem 4(ii)) and maximum error bound (Theorem 4(i)). While the non-convex penalized log-likelihood objective function results in  a non-convex optimization problem, Theorems 2 and 5 specify conditions under which it becomes a convex optimization problem. These conditions favor log-sum penalty over SCAD. Numerical results based on synthetic and real data were presented to illustrate the proposed approaches. In the synthetic data examples the log-sum penalized objective function significantly outperformed the lasso penalized as well as SCAD penalized objective functions with $F_1$-score and Hamming distance as performance metrics.

\appendices
\section{Lemma 2 and Proof of Theorems 1} \label{append1}
Lemma 2 follows from \cite[Lemma 1]{Ravikumar2011} and \cite[Lemma 1]{Tugnait2024}. \\
{\it Lemma 2}. Suppose $\hat{\bm \Sigma} = (1/n) \sum_{t=1}^n {\bm x}(t) {\bm x}^\top (t)$, given $n$ i.i.d.\ samples $\{ {\bm x}(t) \}_{t=1}^n$ of zero-mean Gaussian ${\bm x} \in \mathbb{R}^{mp}$ with covariance ${\bm \Sigma}^\ast$ such that each component $x_i/\sqrt{\Sigma_{ii}^\ast}$ is Gaussian with unit variance. Define 
$\sigma_{max} = \max_{1 \le i \le mp_n} \Sigma_{ii}^\ast$ and 
\begin{equation}  \label{aeqn300}
 \tilde{C}_0 = 40 m \sigma_{max}  \sqrt{2 \ln(4m^2 p_n^\tau)/ \ln(p_n) }.
\end{equation}
Then for any $\tau > 2$, 
\begin{equation}  \label{aeqn305}
   P \Big(  \| \bm{\mathcal C}(\hat{\bm \Sigma}- {\bm \Sigma}^\ast) \|_\infty
	    > \tilde{C}_0 \sqrt{\ln(p_n) / n} \Big) \le 1/p_n^{\tau -2} 
\end{equation}
and
\begin{equation}  \label{aeqn305a}
   P \Big(  \| \hat{\bm \Sigma}- {\bm \Sigma}^\ast) \|_\infty
	    > {C}_0 \sqrt{\ln(p_n) / n} \Big) \le 1/p_n^{\tau -2} 
\end{equation}
if $n > 2 \ln(4m^2 p_n^\tau)$, where $C_0 = \tilde{C}_0/m$. $\quad \bullet$ \\
{\it Proof}. The bound (\ref{aeqn305}) follows from \cite[Lemma 1]{Tugnait2024} when \cite[Lemma 1]{Tugnait2024} is applied to Gaussian distributions. Both \cite[Lemma 1]{Ravikumar2011} and \cite[Lemma 1]{Tugnait2024} are stated for sub-Gaussian distributions, and the latter is based on the former. The bound (\ref{aeqn305a}) follows similar to (\ref{aeqn305}). $\quad \blacksquare$

{\it Proof of Theorem 1}. Let $\bm{\Omega} = \bm{\Omega}^\ast + \bm{\Delta}$ with both $\bm{\Omega}, \, \bm{\Omega}^\ast \succ {\bm 0}$, and 
\begin{equation}
  Q(\bm{\Omega}) := \bar{\cal L} ({\bm \Omega}) - \bar{\cal L} ({\bm \Omega}^\ast) \, .
\end{equation}
The estimate $\hat{\bm{\Omega}}$ minimizes $Q(\bm{\Omega})$, or equivalently, $\hat{\bm{\Delta}} = \hat{\bm{\Omega}} - \bm{\Omega}^\ast$ minimizes $G(\bm{\Delta}) := Q(\bm{\Omega}^\ast + \bm{\Delta})$.
We will follow the method of proof of \cite[Theorem 1]{Tugnait21a}, which, in turn, for the most part, follows the method of proof of \cite[Theorem 1]{Rothman2008} pertaining to lasso penalty. Consider the set
\begin{equation}  \label{naeq1001}
  \Theta_n(R) :=  \left\{ \bm{\Delta} \, :\, 
	 \bm{\Delta} = \bm{\Delta}^\top, \; \|\bm{\Delta} \|_F = R r_n \right\}
\end{equation}
where $R$ and $r_n$ are as in (\ref{neq15ab0}) and (\ref{neq15ab1}), respectively.  Since $G(\hat{\bm{\Delta}}) \le G(\bm{0}) = 0 $, if we can show that $\inf_{\bm{\Delta}}  \{ G(\bm{\Delta}) \, :\, \bm{\Delta} \in \Theta_n(R) \} > 0$, then the minimizer $\hat{\bm{\Delta}}$ must be inside $\Theta_n(R)$, and hence $\| \hat{\bm{\Delta}} \|_F \le   R r_n$.
It is shown in \cite[(9)]{Rothman2008} that
\begin{equation} \label{nneq100}
     \ln (|\bm{\Omega^\ast + \Delta}|) - \ln (|\bm{\Omega}^\ast|) = \mbox{tr} (\bm{\Sigma^\ast  \Delta}) - A_1
\end{equation}
where, with $\bm{H}(\bm{\Omega}^\ast, \bm{\Delta}, v ) = (\bm{\Omega}^\ast+v \bm{\Delta})^{-1} \otimes (\bm{\Omega}^\ast+v \bm{\Delta})^{-1}$ and $v$ denoting a scalar,
\begin{align} \label{naeq1110}
     A_1 = & \mbox{vec}(\bm{\Delta})^\top \left( \int_0^1 (1-v) 
			  \bm{H}(\bm{\Omega}^\ast, \bm{\Delta}, v ) \, dv \right)  \mbox{vec}(\bm{\Delta}) \, .
\end{align}
Noting that $\bm{\Omega}^{-1} = \bm{\Sigma}$, we can rewrite $G({\bm{\Delta}})$ as 
\begin{equation}
     G({\bm{\Delta}}) = A_1 + A_2 + A_3 +A_4\, ,  \label{mainG}
\end{equation}
where
\begin{align} 
    A_2 = & \mbox{tr} \left( (\hat{\bm{\Sigma}} 
		 - \bm{\Sigma}^\ast ) \bm{\Delta}  \right) \, ,  \label{naeq1100} \\
     A_3 = & \alpha  \sum_{i,j=1; i \ne j}^{mp_n} \left( \rho_\lambda (| \Omega_{ij}^\ast +\Delta_{ij} |)
		    - \rho_\lambda (| \Omega_{ij}^\ast  |) \right) \, , \label{naeq1120} \\
		 A_4 = & (1-\alpha) m \sum_{q,\ell=1; q \ne \ell}^{p_n} 
		  \Big( \rho_\lambda ( \|({\bm \Omega}^\ast)^{(q \ell)} + {\bm \Delta}^{(q \ell)}\|_F  )  \nonumber \\
		        & \;\;  - \rho_\lambda ( \|({\bm \Omega}^\ast)^{(q \ell)} \|_F ) \Big) \, . \label{naeq1121}
\end{align}
Following \cite[p.\ 502]{Rothman2008}, we have
\begin{equation}
     A_1  \ge \frac{ \| \bm{\Delta} \|_F^2 }{  2 (\| \bm{\Omega}^\ast \| +  \| \bm{\Delta} \|)^2 }
		  \ge \frac{ \| \bm{\Delta} \|_F^2 }{  2 \left( \beta_{\min}^{-1} + R r_n \right)^2 } \label{boundA1}
\end{equation}
where we have used the fact that $\| \bm{\Omega}^\ast \| = \| (\bm{\Sigma}^\ast)^{-1} \| = \phi_{\max }((\bm{\Sigma}^\ast)^{-1}) = (\phi_{\min }((\bm{\Sigma}^\ast))^{-1} \le \beta_{\min}^{-1} $ and $\| \bm{\Delta} \| \le \| \bm{\Delta} \|_F = R r_n$. We now consider $A_2$ in (\ref{naeq1100}). We have
\begin{equation}
     A_2  =  \underbrace{\sum_{i,j=1; i\ne j}^{mp_n} [\hat{\bm{\Sigma}} - \bm{\Sigma}^\ast]_{ij} \Delta_{ji}}_{L_1} + 
		\underbrace{\sum_{i=1}^{m p_n} [\hat{\bm{\Sigma}} - \bm{\Sigma}^\ast]_{ii} \Delta_{ii} }_{L_2}
\end{equation}
By Lemma 2, the sample covariance $\hat{\bm{\Sigma}}$ satisfies the tail bound
\begin{equation}  \label{naeq58bx}
   P \left(  \max_{k, \ell} \Big| [ \hat{\bm{\Sigma}} - \bm{\Sigma}^\ast ]_{kl} \Big| 
	    > C_0 \sqrt{\frac{\ln( p_n)}{n}} \right) \le \frac{1}{(p_n)^{\tau -2}} 
\end{equation}
for $\tau > 2$, if the sample size  $n >  N_1$ ($N_1$ is defined in (\ref{neq15ab2})). To bound $L_1$, using Lemma 2, with probability $> 1- 1/p_n^{\tau-2}$,
\begin{align}
    | L_1 | & \le  \| \bm{\Delta}^-\|_1 \, \max_{i,j} \big|  [\hat{\bm{\Sigma}} - \bm{\Sigma}^\ast]_{ij}  \big| 
		    \le   \| \bm{\Delta}^-\|_1 \, C_0 \sqrt{\frac{\ln(p_n)}{n}}  \, . \label{naeq1205}
\end{align}
Similarly, by Cauchy-Schwarz inequality, Lemma 2 and (\ref{neq15ab1}), 
\begin{align}
	| L_2 | & \le  \| \bm{\Delta}^+\|_1 \, C_0 \sqrt{\frac{\ln(p_n)}{n}} \nonumber \\
	   &  \le  C_0 \sqrt{\frac{\ln(p_n)}{n}} \, \sqrt{ m p_n} \,  \| \bm{\Delta}^+ \|_F 
				\le \| \bm{\Delta}^+ \|_F C_0 r_n  \, .
\end{align}
Therefore, with probability $> 1- 1/p_n^{\tau-2}$, 
\begin{align}
     | A_2 | & \le \| \bm{\Delta}^-\|_1 \, C_0 \sqrt{\frac{\ln(p_n)}{n}} +
			  \| \bm{\Delta}^+ \|_F C_0 r_n  \, . \label{naeq1330} 
\end{align}
We now derive a different bound on $A_2$. Define $\tilde{\bm{\Delta}} \in \mathbb{R}^{p_n \times p_n}$ with $(i,j)$-th element $\tilde{\Delta}_{ij} = \| \bm{\Delta}^{(ij)} \|_F$, where $\bm{\Delta}^{(ij)} $ is defined from ${\bm \Delta}$ similar to (\ref{addn020}). By Cauchy-Schwarz inequality,
\begin{align}
   \| \bm{\Delta}^-\|_1 = &  \sum_{i,j=1; i\ne j}^{mp_n} |\Delta_{ij}| \le m \| \tilde{\bm{\Delta}}^-\|_1  \nonumber \\
	  & \;  + \underbrace{\Big( \sum_{k=1}^{p_n} \|\bm{\Delta}^{(kk)}\|_1 - \| \bm{\Delta}^+\|_1 \Big)}_{=: B} \, .
\end{align}
Then using $\sum_k \|\bm{\Delta}^{(kk)}\|_1 \le m \sum_k \tilde{\Delta}_{kk} \le m \sqrt{p_n} \, \| \tilde{\bm{\Delta}}^+ \|_F$, we have
\begin{align*}
     |L_2| + C_0 \sqrt{\frac{\ln(p_n)}{n}} \, B & 
		 \le C_0 \sqrt{\frac{\ln(p_n)}{n}} \, (\sum_{k=1}^{p_n} \|\bm{\Delta}^{(kk)}\|_1) \nonumber \\
		& \;\;  \le \| \tilde{\bm{\Delta}}^+ \|_F \sqrt{m}\,  C_0 r_n \, .
\end{align*}
Therefore, an alternative bound is
\begin{align}
     | A_2 | & 
				 \le m \| \tilde{\bm{\Delta}}^-\|_1 \, C_0 \sqrt{\frac{\ln(p_n)}{n}} + \sqrt{m} \,
			  \| \tilde{\bm{\Delta}}^+ \|_F C_0 r_n \, . \label{naeq1330a}
\end{align}

For the rest of the proof we have two different approaches, one for lasso and log-sum and the other for SCAD penalty. 

{\it For Lasso and Log-Sum Penalties}: 
We now bound $A_3$ in (\ref{naeq1120}). Let $\bar{\cal E}^\ast$ denote the true enlarged edge-set corresponding to ${\cal E}^\ast$ when one interprets multi-attribute model as a single-attribute model. Let $(\bar{\cal E}^\ast)^c$ denote its complement. Using the mean-value theorem, we have ($\rho_\lambda^\prime(u) = \frac{d \rho_\lambda (u)}{du}$)
\begin{align}
  \rho_\lambda (| \Omega_{ij}^\ast +\Delta_{ij} |) = & \rho_\lambda (| \Omega_{ij}^\ast |) \nonumber \\
	& +  
	  \rho_\lambda^\prime (| \tilde{\Omega}_{ij} |)  (| \Omega_{ij}^\ast +\Delta_{ij} | - | \Omega_{ij}^\ast  |)
		  \label{expan}
\end{align}
where $|\tilde{\Omega}_{ij}| = |\Omega_{ij}^\ast| + \gamma (| \Omega_{ij}^\ast +\Delta_{ij} | - | \Omega_{ij}^\ast  |)$ for some $\gamma \in [0,1]$. We have   
\begin{align}
   A_3  = & \alpha  \sum_{(i,j) \in \bar{\cal E}^\ast} \rho_\lambda^\prime 
	      (| \tilde{\Omega}_{ij} |)  (| \Omega_{ij}^\ast +\Delta_{ij} | - | \Omega_{ij}^\ast  |) \nonumber \\
			& \;\; 	+ \alpha  \sum_{(i,j) \in (\bar{\cal E}^\ast)^c} \rho_\lambda (| \Delta_{ij} |) \label{eqnn20} \\
	   \ge & - \alpha  \sum_{(i,j) \in \bar{\cal E}^\ast} \rho_\lambda^\prime 
	      (| \tilde{\Omega}_{ij} |)  | \Delta_{ij} |
				+ \alpha  \sum_{(i,j) \in (\bar{\cal E}^\ast)^c} C_\lambda | \Delta_{ij} | \nonumber \\
				& \quad \;\; \mbox{for } \;
				  | \Delta_{ij} | \le \delta_\lambda ,  \label{eqnn2}
\end{align}
using the triangle inequality and (\ref{prop7}) in the last step above. Now use property (viii) of the penalty functions and $C_\lambda = \lambda/2$ to conclude that 
\begin{align}				
A_3 & \ge  - \alpha \lambda_n \sum_{(i,j) \in \bar{\cal E}^\ast} | \Delta_{ij} | 
          + \alpha (\lambda_n/2) \sum_{(i,j) \in (\bar{\cal E}^\ast)^c}  | \Delta_{ij} | \, . \label{naeq1220}
\end{align}
Next we bound $A_4$ in (\ref{naeq1121}). Considering the true edge-set ${\cal E}^\ast$ for the multi-attribute graph, let $({\cal E}^\ast)^c$ denote its complement. If the edge $\{i,j\} \in ({\cal E}^\ast)^c$, then $(\|{\bm \Omega}^\ast)^{(ij)} = {\bm 0}$, therefore, $(\|{\bm \Omega}^\ast)^{(ij)} + {\bm \Delta}^{(ij)}\|_F - \|({\bm \Omega}^\ast)^{(ij)} \|_F = \|{\bm \Delta}^{(ij)}\|_F $. For $\{i,j\} \in {\cal E}^\ast$, by the triangle inequality,
   $\|({\bm \Omega}^\ast)^{(ij)} + {\bm \Delta}^{(ij)}\|_F - \|({\bm \Omega}^\ast)^{(ij)} \|_F 
	  \ge  - \| {\bm \Delta}^{(ij)}\|_F$.
Thus, mimicking the steps for bounding $A_3$, we have
\begin{align}
     A_4 & \ge  - (1-\alpha)m \lambda_n \sum_{(i,j) \in {\cal E}^\ast} \| {\bm \Delta}^{(ij)} \|_F \nonumber \\
      & \;\;     + (1-\alpha)m (\lambda_n/2) \sum_{(i,j) \in (({\cal E}^\ast)^c}  \| {\bm \Delta}^{(ij)} \|_F \, .\label{naeq1220b}
\end{align}

Split $A_2$ as $A_2 = \alpha A_2 + (1-\alpha) A_2$, apply bound (\ref{naeq1330}) to $\alpha A_2$ and (\ref{naeq1330a}) to $(1-\alpha) A_2$, use $\| \bm{\Delta}^- \|_1 = \| \bm{\Delta}^-_{\bar{\cal E}^\ast} \|_1 + \|\bm{\Delta}^-_{(\bar{\cal E}^\ast)^c} \|_1$ and $\| \tilde{\bm{\Delta}}^- \|_1 = \| \tilde{\bm{\Delta}}^-_{{\cal E}^\ast} \|_1 + \|\tilde{\bm{\Delta}}^-_{({\cal E}^\ast)^c} \|_1$. Define $d_1 =\frac{\ln(p_n)}{n}$, then $r_n = \sqrt{mp_n + m^2 s_{n}^\ast} \; d_1$.  We have 
\begin{align}
  \alpha A_2+ & A_3 \ge  - \alpha |A_2| 
	 + \alpha \lambda_n ( 0.5 \|\bm{\Delta}^-_{(\bar{\cal E}^\ast)^c} \|_1 - \| \bm{\Delta}^-_{\bar{\cal E}^\ast} \|_1 ) 
	    \nonumber \\
	\ge & \alpha ( 0.5 \lambda_n - C_0 d_1) \sum_{(i,j) \in (\bar{\cal E}^\ast)^c}  | \Delta_{ij} |  \nonumber \\
	&  - \alpha (  \lambda_n + C_0 d_1) \sum_{(i,j) \in \bar{\cal E}^\ast}  | \Delta_{ij} |
	 - \alpha C_0 r_n \| \bm{\Delta}^+ \|_F \, .
\end{align}
Since we pick $\lambda_n \ge \lambda_{n\ell}$ in Theorem 1, $0.5 \lambda_n - C_0 d_1 \ge 0$ and therefore, the first term above can be neglected. Now $\sum_{(i,j) \in \bar{\cal E}^\ast}  | \Delta_{ij} | \le \sqrt{m^2 s_{n}^\ast} \|{\bm \Delta}\|_F$, by the Cauchy-Schwarz inequality, and $\| \bm{\Delta}^+ \|_F \le \| \bm{\Delta} \|_F$. We then have
\begin{align}	
	\alpha A_2+ & A_3 \ge  - \alpha \left( (  \lambda_n + C_0 d_1) \sqrt{m^2 s_{n}^\ast} 
	           + C_0 r_n \right) \| \bm{\Delta} \|_F \, .
	\label{naeq5200}
\end{align}
Similarly, we have
\begin{align}	
	(1-\alpha) A_2+ & A_4 \ge  -(1- \alpha)m \nonumber \\
	  & \;\; \times \left( (  \lambda_n + C_0 d_1) \sqrt{ s_{n}^\ast} 
	           + C_0 r_n \right) \| \bm{\Delta} \|_F \, .
	\label{naeq5210}
\end{align}
From (\ref{naeq5200}) and (\ref{naeq5210}) we have
\begin{align}
  A_2+ & A_3+A_4 \ge  - \| \bm{\Delta} \|_F  \left( \lambda_n m \sqrt{s_{n}^\ast} + C_0 d_1 m \sqrt{s_{n}^\ast}
	      + m C_0 r_n \right) \nonumber \\
	  & \ge - \| \bm{\Delta} \|_F  \left( \lambda_n m \sqrt{s_{n}^\ast} + (1+ m) C_0 r_n \right)	\nonumber \\
	  &
		 \ge - 2   (1+ m) C_0 r_n \| \bm{\Delta} \|_F \label{naeq5330}
\end{align}
where we used the fact that since $\lambda_n \le \lambda_{nu1}$, $ \lambda_n m \sqrt{s_{n}^\ast} \le C_0 (1+m) r_n$.
Using (\ref{mainG}), the bound (\ref{boundA1}) on $A_1$, bound (\ref{naeq5330}) on $A_2+  A_3+A_4$, and $\| \bm{\Delta} \|_F = R r_n$, we have with probability $> 1- 1/p_n^{\tau-2}$, 
\begin{align}
     G({\bm{\Delta}}) \ge &  \; \| \bm{\Delta} \|_F^2 \left[ \frac{1}{2 (\beta_{\min}^{-1} + R r_n )^2}  
		     - \frac{2C_0(1+m)}{R} \right]  \, .\label{naeq1370} 
\end{align}
For the given choice of $N_2$, $R r_n \le R r_{N_2} \le 0.1/\beta_{\min}$ for $n \ge N_2$. Also, $2C_0(1+m)/R = \beta_{\min}^2/4$ by (\ref{neq15ab0}). Then for $n \ge N_2$,
\begin{align*}
    \frac{1}{2 (\beta_{\min}^{-1} + R r_n )^2}- \frac{2C_0(1+m)}{R}
		\ge \beta_{\min}^2 \left( \frac{1}{2.42} - \frac{1}{4} \right) > 0 \, ,
\end{align*}
implying $G({\bm{\Delta}})  > 0$. This proves (\ref{neq15}).  The choice of $N_3$ for log-sum penalty ensures that $| \Delta_{ij} | \le \delta_\lambda = \epsilon$  needed in (\ref{eqnn2}) is satisfied w.h.p.: if $R r_n \le  \epsilon$, then  $| \Delta_{ij} | \le  \|{\bm \Delta}\|_F \le R r_n \le \epsilon$. 

{\it For SCAD Penalty}:
Here we address (\ref{expan}) differently. Using triangle inequality, we have 
\begin{align}
  |\tilde{\Omega}_{ij}| & \ge |\Omega_{ij}^\ast| + \gamma \big( | \Omega_{ij}^\ast  | - | \Delta_{ij} | - | \Omega_{ij}^\ast  | \big) \nonumber \\
	  & \ge |\Omega_{ij}^\ast| - | \Delta_{ij} | \, . \label{expan2}
\end{align}
Since $| \Delta_{ij} | \le \| \bm{\Delta} \|_F \le R r_n$, the choice $\lambda_n = \lambda_{nu2}$ implies that $\lambda_n \ge Rr_n$, satisfying $| \Delta_{ij} | \le  \lambda_n$. Therefore, $|\tilde{\Omega}_{ij}| \ge |\Omega_{ij}^\ast| - \lambda_n$. 
For $n \ge N_4$, $\rho_\lambda^\prime (|\tilde{\Omega}_{ij}|)=0$ (see (\ref{neq15ab41})) if $\{i,j\} \in \bar{\cal E}^\ast$, i.e, $| \Omega_{ij}^\ast  | \ne 0$, since in this case $|\tilde{\Omega}_{ij}| \ge (a+1) \lambda_n  - \lambda_n = a \lambda_n$. As in (\ref{eqnn20}), we have
\begin{align}
   A_3  = & \, \alpha  \sum_{(i,j) \in \bar{\cal E}^\ast} \rho_\lambda^\prime 
	      (| \tilde{\Omega}_{ij} |)  (| \Omega_{ij}^\ast +\Delta_{ij} | - | \Omega_{ij}^\ast  |) \nonumber \\
			& \;\; 	+ \alpha  \sum_{(i,j) \in (\bar{\cal E}^\ast)^c} \rho_\lambda (| \Delta_{ij} |) \nonumber \\
	   \ge & \, \alpha  \sum_{(i,j) \in (\bar{\cal E}^\ast)^c} C_\lambda | \Delta_{ij} |
				 \quad  \mbox{for } \;
				  | \Delta_{ij} | \le \delta_\lambda ,  \label{eqnn2a} \\
		  = & \, 
		\alpha (\lambda_n/2) \sum_{(i,j) \in (\bar{\cal E}^\ast)^c}  | \Delta_{ij} | \, . \label{eqnn22}
\end{align} 
Mimicking the steps for bounding $A_3$ above and under same conditions, we have
\begin{align}
     A_4 & \ge   (1-\alpha)m (\lambda_n/2) \sum_{(i,j) \in ({\cal E}^\ast)^c}  \| {\bm \Delta}^{(ij)} \|_F \, .\label{naeq1220b2}
\end{align} 
Thus
\begin{align}
  \alpha A_2+ & A_3 \ge  - \alpha |A_2| 
	 + 0.5 \, \alpha \lambda_n  \|\bm{\Delta}^-_{(\bar{\cal E}^\ast)^c} \|_1  
	    \nonumber \\
	\ge & \, \alpha ( 0.5 \lambda_n - C_0 d_1) \sum_{(i,j) \in (\bar{\cal E}^\ast)^c}  | \Delta_{ij} |  \nonumber \\
	&  - \alpha C_0 d_1 \sum_{(i,j) \in \bar{\cal E}^\ast}  | \Delta_{ij} |
	 - \alpha C_0 r_n \| \bm{\Delta}^+ \|_F \, .
\end{align}
Since we pick $\lambda_n = \max \left( R r_n , \lambda_{n u1} \right)$ in Theorem 1, $0.5 \lambda_n - C_0 d_1 \ge 0$ and therefore, the first term above can be neglected. Now $\sum_{(i,j) \in \bar{\cal E}^\ast}  | \Delta_{ij} | \le \sqrt{m^2 s_{n}^\ast} \|{\bm \Delta}\|_F$, by the Cauchy-Schwarz inequality, and $\| \bm{\Delta}^+ \|_F \le \| \bm{\Delta} \|_F$. We then have
\begin{align}	
	\alpha A_2+ & A_3 \ge  - \alpha \left( C_0 d_1 \sqrt{m^2 s_{n}^\ast} 
	           + C_0 r_n \right) \| \bm{\Delta} \|_F \, .
	\label{naeq52002}
\end{align}
By very similar arguments we also have
\begin{align}	
	(1-\alpha) A_2+ & A_4 \ge  -(1- \alpha)m \nonumber \\
	  & \;\; \times \left( C_0 d_1 \sqrt{ s_{n}^\ast} 
	           +  C_0 r_n \right) \| \bm{\Delta} \|_F \, .
	\label{naeq52102}
\end{align}
From (\ref{naeq52002}) and (\ref{naeq52102}) we have
\begin{align}
  A_2+ & A_3+A_4 \ge  - \| \bm{\Delta} \|_F  \left( C_0 d_1 m \sqrt{s_{n}^\ast}
	      + m C_0 r_n \right) \nonumber \\
	  &
		 \ge -    (1+ m) C_0 r_n \| \bm{\Delta} \|_F \label{naeq5330}
\end{align}
where we used the fact that $C_0 d_1 m \sqrt{s_{n}^\ast} \le C_0 r_n$.
Mimicking (\ref{naeq1370}), with probability $> 1- 1/p_n^{\tau-2}$, we have
\begin{align}
     G({\bm{\Delta}}) \ge &  \; \| \bm{\Delta} \|_F^2 \left[ \frac{1}{2 (\beta_{\min}^{-1} + R r_n )^2}  
		     - \frac{(1+m) C_0 }{R} \right]  \nonumber \\
				\ge& \beta_{\min}^2 \left( \frac{1}{2.42} - \frac{1}{8} \right) > 0 \, , \label{naeq13702} 
\end{align}
implying $G({\bm{\Delta}})  > 0$. This proves (\ref{neq15}). For the SCAD penalty, we  need $| \Delta_{ij} | \le \delta_\lambda = \lambda_n$ in (\ref{eqnn22}). Since $| \Delta_{ij} | \le \| \bm{\Delta} \|_F \le R r_n$, the choice $\lambda_n = \lambda_{nu2}$ implies that $\lambda_n \ge Rr_n$, satisfying $| \Delta_{ij} | \le  \lambda_n$. This completes the proof.   $\quad \blacksquare$

\section{Proofs of Lemma 1 and Theorems 2 and 3} \label{append2}

{\it Proof of Lemma 1}. Consider $h({\bm \Omega}) = {\cal L}({\bm \Omega})- \frac{\mu}{2} \| {\bm \Omega} \|_F^2$ for some $\mu \ge 0$. The Hessian of ${\cal L}({\bm \Omega})$ w.r.t.\ $\mbox{vec}({\bm \Omega})$ is $\nabla^2 {\cal L}({\bm \Omega}) = {\bm \Omega}^{-1} \otimes {\bm \Omega}^{-1}$ 
with 
\begin{align}
     \phi_{\min}(\nabla^2 {\cal L}({\bm \Omega})) =  &  \phi_{\min}^2( {\bm \Omega}^{-1})
		  =  1/ \phi_{\max}^2( {\bm \Omega})   = 1 / \| {\bm \Omega} \|^2 .
			 \label{naeq8372} 
\end{align}
Since $\nabla^2 h({\bm \Omega}) = {\bm \Omega}^{-1} \otimes {\bm \Omega}^{-1} - \mu {\bm I}_{(mp)^2}$, it follows that $h({\bm \Omega})$  is positive semi-definite, hence convex, if  
\begin{align}
      \| {\bm \Omega} \| & \le \sqrt{\frac{1}{\mu}} \, .
			 \label{naeq8375} 
\end{align}
By property (v) of the penalty functions, $g(u):=\rho_\lambda(u) +\frac{\mu}{2} u^2$ is convex, for some $\mu \ge 0$, and by property (ii), it is non-decreasing on $\mathbb{R}_+$. Therefore, by the composition rules \cite[Sec.\ 3.2.4]{Boyd2004}, $g(| [ {\bm{\Omega}} ]_{ij}|)$ and $g(\| {\bm \Omega}^{(q \ell )} \|_F)$ are convex. Hence,
\begin{align}
       P_e({\bm \Omega}) + \frac{\mu_e}{2} \| {\bm \Omega} \|_F^2  
		 = &   \sum_{i \ne j}^{mp_n} 
		 \Big( \rho_\lambda(\big| [ {\bm{\Omega}} ]_{ij} \big|) +\frac{\mu_e}{2} \big| [ {\bm{\Omega}} ]_{ij} \big|^2 \Big)	 
		  \label{naeq8378}
\end{align}
is convex for $\mu_e = \mu \ge 0$, and similarly,
\begin{align}
  P_g({\bm \Omega}) + \frac{\mu_g}{2} \| {\bm \Omega} \|_F^2  
		 = & m   \; \sum_{ q \ne \ell}^{p_n} \; 
		   \Big( \rho_\lambda ( \| {\bm \Omega}^{(q \ell )} \|_F )
			+\frac{\mu_g}{2 m } \| {\bm \Omega}^{(q \ell )} \|_F^2 \Big)	
				  \label{naeq8379}
\end{align}
is convex for $\mu_g = m  \, \mu$, where $\mu$ is the value that renders $\rho_\lambda(u) +\frac{\mu}{2} u^2$ convex. Express $\bar{\cal L}({\bm \Omega})$ as
\begin{align}
   \bar{\cal L}({\bm \Omega}) = & \alpha \bar{\cal L}_e({\bm \Omega})  
	 + (1-\alpha) \bar{\cal L}_g({\bm \Omega}) \, , \label{naeq8381} \\
	\bar{\cal L}_e({\bm \Omega}) = & 	{\cal L}({\bm \Omega}) - \frac{\mu}{2} \| {\bm \Omega} \|_F^2
	  + P_e({\bm \Omega}) + \frac{\mu}{2} \| {\bm \Omega} \|_F^2  \, , \label{naeq8381a} \\
		\bar{\cal L}_g({\bm \Omega}) = & {\cal L}({\bm \Omega}) - \frac{\mu}{2} \| {\bm \Omega} \|_F^2
	  + P_g({\bm \Omega}) + \frac{\mu}{2} \| {\bm \Omega} \|_F^2   \, .
				  \label{naeq8381b}
\end{align}
Now $\bar{\cal L}_e({\bm \Omega})$ is convex function of ${\bm \Omega}$ if
$\| {\bm \Omega} \|  \le \sqrt{\frac{1}{\mu}}$, 
and $\bar{\cal L}_g({\bm \Omega})$ is convex in ${\bm \Omega}$ if
$\| {\bm \Omega} \|  \le \sqrt{\frac{1}{\mu_g}} = \sqrt{\frac{1}{m \mu }}$.
Thus, for $\bar{\cal L}({\bm \Omega})$ to be strictly convex, using the (minimum) values of $\mu$ to make $\rho_\lambda(u) +\frac{\mu}{2} u^2$ convex, we require
\begin{align}  
		   \| \bm{\Omega} \| < & \;\; \bar{\mu}   \nonumber \\
		 = & \left\{ \begin{array}{ll}
		   \infty & : \;\; \mbox{lasso} \\
			  \sqrt{(a-1)/m } & : \;\; \mbox{SCAD} \\
				\sqrt{  \epsilon /(m \lambda_n) } & : \;\; \mbox{log-sum}, \end{array} \right. \label{naeq8390} 
\end{align}
The choice $\| \bm{\Omega} \| < \bar{\mu} $ makes ${\cal L}({\bm \Omega})- \frac{\mu}{2} \| {\bm \Omega} \|_F^2$ positive definite, hence strictly convex. We take $\| \bm{\Omega} \| = 0.99 \,  \bar{\mu}$, completing the proof.  $\quad \blacksquare$

{\it Proof of Theorem 2}. If $1/\beta_{\min} \le 0.99 \, \bar{\mu}$, then $\bm{\Omega}^\ast \in {\cal B}$ since $\| \bm{\Omega}^\ast \| \le 1/\beta_{\min}$ by assumption (A2). Now we establish that $\hat{\bm{\Omega}}  \in {\cal B}$. To this end, consider 
\begin{align}  
		   \| \hat{\bm{\Omega}} \| \le & \| \hat{\bm{\Omega}} - \bm{\Omega}^\ast \|
			  + \| \bm{\Omega}^\ast \|  \nonumber \\
				 \le & \| \hat{\bm{\Omega}} - \bm{\Omega}^\ast \|_F
			  + \| \bm{\Omega}^\ast \|  \nonumber \\
		 \le & R r_n + 1/\beta_{\min} \, .
		\label{naeq8394} 
\end{align}
Therefore, $\hat{\bm{\Omega}}  \in {\cal B}$. Thus, both  $\hat{\bm{\Omega}}$ and ${\bm{\Omega}}^\ast$ are feasible. The desired result then follows from Theorem 1 and (local) strict convexity of $\bar{\cal L}({\bm \Omega})$ over ${\cal B}$ implied by Lemma 1. $\quad \blacksquare$

{\it Proof of Theorem 3}. We have $\| \hat{\bm \Omega}^{(q \ell)} - ({\bm \Omega}^\ast)^{(q \ell)} \|_F \le \| \hat{\bm \Omega} - {\bm \Omega}^\ast \|_F \le \bar{\sigma}_n$ w.h.p. For the edge $\{ q,\ell \} \in {\cal E}^\ast$, we have
\begin{align}  
		   \| \hat{\bm \Omega}^{(q \ell )} \|_F = & 
			    \| ({\bm \Omega}^\ast)^{(q \ell )} + \hat{\bm \Omega}^{(q \ell)} - ({\bm \Omega}^\ast)^{(q \ell )} \|_F  \nonumber \\
		 \ge & \| ({\bm \Omega}^\ast)^{(q \ell )} \|_F - \| \hat{\bm \Omega}^{(q \ell )} - ({\bm \Omega}^\ast)^{(q \ell )} \|_F \nonumber \\
		 \ge & \nu - \bar{\sigma}_n \ge 0.6 \, \nu \;\; \mbox{ for } \;\; n \ge N_4 \nonumber \\
		 > & \theta_n \, .
		\label{naeq8397} 
\end{align}
Thus, ${\cal E}^\ast \subseteq \hat{\cal E}$. Now consider the set complements $({\cal E}^\ast)^c$ and $\hat{\cal E}^c$. For the edge $\{ q,\ell \} \in ({\cal E}^\ast)^c$, $\| ({\bm \Omega}^\ast)^{(q \ell )} \|_F = 0$. For $n \ge N_4$, w.h.p.\ we have
\begin{align}  
		   \| \hat{\bm \Omega}^{(q \ell)} \|_F \le & 
			 \| ({\bm \Omega}^\ast)^{(q \ell )} \|_F + \| \hat{\bm \Omega}^{(q \ell)} - ({\bm \Omega}^\ast)^{(q \ell)} \|_F \nonumber \\
		 \le & 0 + \bar{\sigma}_n \le 0.4 \, \nu <  \theta_n \, ,
		\label{naeq8399} 
\end{align}
implying that $\{ q,\ell \} \in (\hat{\cal E}^\ast)^c$. Thus, $({\cal E}^\ast)^c \subseteq \hat{\cal E}^c$, hence $\hat{\cal E} \subseteq {\cal E}^\ast$, establishing $\hat{\cal E} = {\cal E}^\ast$.
 $\quad \blacksquare$

\section{Technical Lemmas and Proofs of Theorems 4 and 5} \label{append3}
In this Appendix, we prove Theorems 4 and 5. A first-order necessary condition for minimization of non-convex $\bar{\cal L}({\bm \Omega})$, given by (\ref{eqth2_20}), w.r.t.\ ${\bm \Omega} \in \mathbb{R}^{mp_n \times mp_n}$ is that the zero matrix belongs to the sub-differential of $\bar{\cal L}({\bm \Omega})$ at the solution $\hat{\bm \Omega}$. That is, at ${\bm \Omega} = \hat{\bm \Omega}$, 
\begin{align} 
  {\bm 0} \, \in & \, \partial \bar{\cal L}({\bm \Omega})
	  = \frac{\partial {\cal L}({\bm \Omega})}{\partial {\bm \Omega}} + \alpha \partial P_e({\bm \Omega})
		  + (1-\alpha) \partial P_g({\bm \Omega}) \nonumber \\ 
		= & \hat{\bm \Sigma} - {\bm \Omega}^{-1}  + \alpha \lambda_n {\bm Z}({\bm \Omega}) 
			+ (1-\alpha)m \lambda_n {\bm Y}({\bm \Omega}) 
  \label{maeqn210}  
\end{align}
where $\lambda_n {\bm Z}({\bm \Omega}) \in \partial \sum_{i \ne j}^{mp_n} \rho_\lambda \big( | {\Omega}_{ij} | \big) \in \mathbb{R}^{mp_n \times mp_n}$, the sub-differential of (possibly non-convex) element-wise penalty term for $i \ne j$, is given by 
\begin{align} 
  [{\bm Z}({\bm \Omega})]_{ij} =& \left\{ \begin{array}{l} 
	  v \in [-1,1],  \mbox{ if } 
				                  {\Omega}_{ij} = 0 \\
	      \frac{{\Omega}_{ij}}{|{\Omega}_{ij}|} \mbox{ if } 
				                  {\Omega}_{ij} \ne 0 \; : \mbox{ lasso}\\
		C_{eij} \mbox{ if } 
				                  {\Omega}_{ij} \ne 0	\; : \mbox{ log-sum}	\\	
		D_{eij} \mbox{ if } 
				                  {\Omega}_{ij} \ne 0	\; : \mbox{ SCAD} \, ,	\end{array} \right. \label{maeqn215} 
\end{align}
\begin{align} 
		C_{eij}  = & \frac{\epsilon}{\epsilon + |{\Omega}_{ij}|} \, 
	      \frac{{\Omega}_{ij}}{|{\Omega}_{ij}|} \, ,
\end{align}
\begin{align} 
		D_{eij} = & \left\{ \begin{array}{l} 
		\frac{{\Omega}_{ij}}{|{\Omega}_{ij}|} \mbox{ if } 
				                  0 < | {\Omega}_{ij}| \le \lambda_n \\
		 \frac{a - |{\Omega}_{ij}|/ \lambda_n}{a-1} 
		  \frac{{\Omega}_{ij}}{|{\Omega}_{ij}|}  \mbox{ if } 
				                  \lambda_n < | {\Omega}_{ij} | \le a \lambda_n \\
							{\bm 0} \mbox{ if } 
				                 a \lambda_n < | {\Omega}_{ij} | \, , \end{array} \right.
\end{align}
and $\lambda_n {\bm Y}({\bm \Omega}) \in m^{-1} \partial \sum_{k \ne \ell}^{p_n} \rho_\lambda \big( \| {\bm \Omega}^{(k \ell)} \|_F \big) \in \mathbb{R}^{mp_n \times mp_n}$, the sub-differential of (possibly non-convex) group penalty term for $k \ne \ell$, is given by 
\begin{align} 
  ({\bm Y}({\bm \Omega}))^{(k \ell)} =& \left\{ \begin{array}{l} 
	  {\bm V} \in \mathbb{R}^{m \times m} , \, \| {\bm V} \|_F \le 1, \\
		\quad\quad \mbox{ if } 
				                  \| {\bm \Omega}^{(k \ell)} \|_F = 0 \\
	      \frac{{\bm \Omega}^{(k \ell)}}{\| {\bm \Omega}^{(k \ell)} \|_F} \mbox{ if } 
				                  \| {\bm \Omega}^{(k \ell)} \|_F \ne 0 \; : \mbox{ lasso}\\
		C_g^{(k \ell)} \mbox{ if } 
				                  \| {\bm \Omega}^{(k \ell)} \|_F \ne 0	\; : \mbox{ log-sum}	\\	
		D_g^{(k \ell)} \mbox{ if } 
				                  \| {\bm \Omega}^{(k \ell)} \|_F \ne 0	\; : \mbox{ SCAD} \, ,	\end{array} \right. \label{maeqn216} 
\end{align}
\begin{align} 
		C_g^{(k \ell)} = & \frac{\epsilon}{\epsilon + \| {\bm \Omega}^{(k \ell)} \|_F} \, 
	      \frac{{\bm \Omega}^{(k \ell)}}{\| {\bm \Omega}^{(k \ell)} \|_F} \, ,
\end{align}
\begin{align} 
		D_g^{(k \ell)} = & \left\{ \begin{array}{l} 
		\frac{{\bm \Omega}^{(k \ell)}}{\| {\bm \Omega}^{(k \ell)} \|_F} \mbox{ if } 
				                  0 < \| {\bm \Omega}^{(k \ell)} \|_F \le \lambda_n \\
		 \frac{a - \| {\bm \Omega}^{(k \ell)} \|_F/ \lambda_n}{a-1} 
		  \frac{{\bm \Omega}^{(k \ell)}}{\|{\bm \Omega}^{(k \ell)}\|_F} \\
			\quad\quad \mbox{ if } 
				                  \lambda_n < \| {\bm \Omega}^{(k \ell)} \|_F \le a \lambda_n \\
							{\bm 0} \mbox{ if } 
				                 a \lambda_n < \| {\bm \Omega}^{(k \ell)} \|_F \, . \end{array} \right.
\end{align}
We have $|[{\bm Z}({\bm \Omega})]_{ij}| \le 1$ and $\|({\bm Y}({\bm \Omega}))^{(k \ell)} \|_F = \| \mbox{vec}(({\bm Y}({\bm \Omega}))^{(k \ell)}) \|_2 \le 1$ for all three penalties; note that $[{\bm Z}({\bm \Omega})]_{ij} = 0$ for $i=j$ and $({\bm Y}({\bm \Omega}))^{(k \ell)} = {\bm 0}$ for $k = \ell$. Suppose that $\hat{\bm \Omega}$ is a solution to (\ref{maeqn210}) which is a first-order necessary condition for a stationary point of $\bar{\cal L}({\bm \Omega})$. Theorem 4 addresses some properties of this $\hat{\bm \Omega}$. 

Let $\tilde{\bm \Omega}$ be a stationary point of $ \bar{\cal L}({\bm \Omega})$ under the constraint ${\bm \Omega}_{S^c} = {\bm 0}$, i.e., $\tilde{\bm \Omega}$ is a solution to ${\bm 0} \, \in  \, \partial \big( \bar{\cal L}({\bm \Omega}) \big|_{{\bm \Omega}_{S^c} ={\bm 0}} \big)$.  Define
\begin{align}  
		  {\bm \Delta} = & \tilde{\bm \Omega} - {\bm \Omega}^\ast  \, , \label{meq1000} \\
		 {\bm R}( {\bm \Delta}) = & \tilde{\bm \Omega}^{-1} - ({\bm \Omega}^\ast)^{-1}
			          + ({\bm \Omega}^\ast)^{-1} {\bm \Delta} ({\bm \Omega}^\ast)^{-1}  \, , \label{meq1010} \\ 
		{\bm W} = & \hat{\bm \Sigma} - {\bm \Sigma}^\ast  \, . \label{meq1020} 
\end{align}
{\it Lemma 3}. If $\| {\bm \Delta}\|_\infty < 1/(3 \bar{\kappa}_{\Sigma^\ast} \bar{d}_n)$ then 
\begin{align}
   \| {\bm R}( {\bm \Delta}) \|_\infty \, \le \, & \frac{3}{2} \bar{d}_n \, \| {\bm \Delta}\|_\infty^2 
	 \bar{\kappa}_{\Sigma^\ast}^3 \, . \label{meq1030}
\end{align}
If $\| \bm{\mathcal C}({\bm \Delta})\|_\infty < 1/(3 {\kappa}_{\Sigma^\ast} {d}_n)$ then 
\begin{align}
   \| \bm{\mathcal C}({\bm R}( {\bm \Delta})) \|_\infty \, \le \,&
	\frac{3}{2} {d}_n \, \| \bm{\mathcal C}({\bm \Delta})\|_\infty^2 
	 {\kappa}_{\Sigma^\ast}^3 \, .  \label{meq1040}
\end{align}
{\it Proof}. The bound (\ref{meq1030}) is proved in \cite[Lemma 5]{Ravikumar2011} and the bound (\ref{meq1040}) is proved in \cite[Lemma 9]{Kolar2014}. $\quad \blacksquare$

Lemma 4 establishes sufficient conditions under which $\tilde{\bm \Omega}$ is also a solution to ${\bm 0} \, \in  \, \partial  \bar{\cal L}({\bm \Omega})$. \\
{\it Lemma 4}. If $\max \big( \|{\bm W}\|_\infty \, , \| {\bm R}( {\bm \Delta}) \|_\infty \big)  \le \gamma \lambda_n /4$ and $\max \big( \| \bm{\mathcal C}({\bm W}) \|_\infty \, , \| \bm{\mathcal C}({\bm R}( {\bm \Delta}) )\|_\infty \big)  \le \gamma m \lambda_n /4$, then  ${\bm 0} \, \in  \, \partial  \bar{\cal L}({\bm \Omega}) \big|_{{\bm \Omega} =\tilde{\bm \Omega}}$ $\quad \bullet$ \\
{\it Proof}. This is a key step in the primal-dual witness approach of \cite{Ravikumar2011} for single-attribute graphs and that of \cite{Kolar2014} for multi-attribute graphs. With 
\begin{align}  
	{\bm X}(\tilde{\bm \Omega}) & = \alpha \lambda_n {\bm Z}(\tilde{\bm \Omega}) 
			+ (1-\alpha)m \lambda_n {\bm Y}(\tilde{\bm \Omega})  \, ,  \label{meq1060}  
\end{align} 
(\ref{maeqn210}) can be expressed as $\hat{\bm \Sigma} - \tilde{\bm \Omega}^{-1}  + {\bm X}(\tilde{\bm \Omega}) = {\bm 0}$. By construction of $\tilde{\bm \Omega}$, $\tilde{\bm \Omega}_{S^c} = {\bm 0}$ and $\hat{\bm \Sigma}_S - (\tilde{\bm \Omega}^{-1})_S  + {\bm X}(\tilde{\bm \Omega}_S) = {\bm 0}$. For the unconstrained problem, we need to show that (\ref{maeqn210}) holds for ${\bm \Omega} =\tilde{\bm \Omega}$, equivalently,
\begin{align} 
   \hat{\bm \Sigma}_S - (\tilde{\bm \Omega}^{-1})_S  + {\bm X}(\tilde{\bm \Omega}_S) = & {\bm 0}  \, ,  \label{meq1050a} \\
	  \hat{\bm \Sigma}_{S^c} - (\tilde{\bm \Omega}^{-1})_{S^c} 
		   + {\bm X}(\tilde{\bm \Omega}_{S^c}) = & {\bm 0}  \, .  \label{meq1050b}
\end{align}
Now (\ref{meq1050a}) is true for the constrained problem. It remains to show that ${\bm X}(\tilde{\bm \Omega}_{S^c}) =  (\tilde{\bm \Omega}^{-1})_{S^c} - \hat{\bm \Sigma}_{S^c}$ is a valid solution to (\ref{meq1050b}) with ${\bm Z}(\tilde{\bm \Omega}_{S^c})$ and ${\bm Y}(\tilde{\bm \Omega}_{S^c})$ satisfying the sub-differential conditions (\ref{maeqn215}) and (\ref{maeqn216}) for every $e_f \in f \in S^c$, so that $\tilde{\bm \Omega}$ qualifies as a stationary point of $\bar{\cal L}({\bm \Omega})$.
To this end, we first rewrite (\ref{maeqn210}) as
\begin{align} 
   \hat{\bm \Sigma} & - ({\bm \Omega}^\ast)^{-1} + ({\bm \Omega}^\ast)^{-1} {\bm \Delta} ({\bm \Omega}^\ast)^{-1}  
	 - {\bm R}( {\bm \Delta}) + {\bm X}(\tilde{\bm \Omega}) = {\bm  0} \, .  \label{meq1050} 
\end{align}
In terms of $m \times m$ submatrices of ${\bm \Delta}$, $\hat{\bm \Sigma}$, ${\bm \Omega}^\ast$ and ${\bm X}(\tilde{\bm \Omega})$ corresponding to various graph edges, using $\mbox{bvec}({\bm A} {\bm D} {\bm B}) = ({\bm B}^\top \boxtimes {\bm A}) \mbox{bvec}({\bm D})$ \cite[Lemma 1]{Tracy1989}, we may rewrite (\ref{meq1050}) as  
\begin{align} 
   {\bm \Gamma}^\ast \mbox{bvec}({\bm \Delta} )  + \mbox{bvec}({\bm W} - {\bm R}( {\bm \Delta})) 
	   + \mbox{bvec}({\bm X}(\tilde{\bm \Omega}))   = {\bm 0} \, ,
  \label{meqa220}  
\end{align}
which then can be rewritten as
\begin{align} 
  &  \begin{bmatrix} {\bm \Gamma}^\ast_{S,S} & {\bm \Gamma}^\ast_{S,S^c} \\
	   {\bm \Gamma}^\ast_{S^c,S} & {\bm \Gamma}^\ast_{S^c,S^c} \end{bmatrix} 
		 \begin{bmatrix} \mbox{bvec}({\bm \Delta}_S ) \\ \mbox{bvec}({\bm \Delta}_{S^c} ) \end{bmatrix}
	-	\begin{bmatrix} \mbox{bvec}(({\bm W}-{\bm R}( {\bm \Delta}))_S) \\ 
	  \mbox{bvec}(({\bm W}-{\bm R}( {\bm \Delta}))_{S^c} ) \end{bmatrix}  \nonumber \\
	& \quad	+  \begin{bmatrix} \mbox{bvec}({\bm X}(\tilde{\bm \Omega}_S)) \\
			   \mbox{bvec}({\bm X}(\tilde{\bm \Omega}_{S^c})) \end{bmatrix} = 
				\begin{bmatrix} {\bm 0} \\ {\bm 0} \end{bmatrix} \, . \label{meqa227}  
\end{align}
Since $\tilde{\bm \Delta}_{S^c} = \tilde{\bm \Omega}_{S^c} - {\bm \Omega}^\ast_{S^c} ={\bm 0}$ by construction, (\ref{meqa227}) reduces to 
\begin{align} 
   & {\bm \Gamma}^\ast_{S,S} \, \mbox{bvec}({\bm \Delta}_S )  + \mbox{bvec}(({\bm W}-{\bm R}( {\bm \Delta}))_S) 
	   \nonumber \\
	 & \quad     + \mbox{bvec}({\bm X}(\tilde{\bm \Omega}_S))     = {\bm 0} \, ,  \label{meq2000a}  \\
	& {\bm \Gamma}^\ast_{S^c,S} \, \mbox{bvec}({\bm \Delta}_S )  + \mbox{bvec}(({\bm W}-{\bm R}( {\bm \Delta}))_{S^c}) 
	 \nonumber \\ & \quad    + \mbox{bvec}({\bm X}(\tilde{\bm \Omega}_{S^c}))    = {\bm 0} \, .  \label{meq2000b}
\end{align}
By construction of $\tilde{\bm \Omega}$ as a stationary point of $\bar{\cal L}({\bm \Omega})$ under the constraint ${\bm \Delta}_{S^c} = {\bm 0}$, (\ref{meq2000a}) is satisfied. It remains to show that (\ref{meq2000b}) is true. Substituting for $\mbox{bvec}({\bm \Delta}_S )$ from (\ref{meq2000a}) into (\ref{meq2000b}), we have 
\begin{align} 
	\mbox{bvec} & ({\bm X}(\tilde{\bm \Omega}_{S^c})) =  {\bm \Gamma}^\ast_{S^c,S} ({\bm \Gamma}^\ast_{S,S})^{-1}
	  \Big( \mbox{bvec}(({\bm W}-{\bm R}( {\bm \Delta}))_S) \nonumber \\
		& + \mbox{bvec}({\bm X}(\tilde{\bm \Omega}_S)) \Big)
		- \mbox{bvec}(({\bm W}-{\bm R})_{S^c})
	 \, .  \label{meq2010}
\end{align}
Using (\ref{meq1060}), we split (\ref{meq2010}) as 
\begin{align} 
 \alpha \lambda_n  &\, \mbox{bvec}({\bm Z}(\tilde{\bm \Omega}_{S^c})) 
	 = \alpha {\bm \Gamma}^\ast_{S^c,S} ({\bm \Gamma}^\ast_{S,S})^{-1} \nonumber \\
	  &  \times \Big( \mbox{bvec}(({\bm W}-{\bm R}( {\bm \Delta}))_S)  
		 + \lambda_n \, \mbox{bvec}({\bm Z}(\tilde{\bm \Omega}_S)) \Big) \nonumber \\
	& - \alpha \, \mbox{bvec}(({\bm W}-{\bm R}( {\bm \Delta}))_{S^c})
	 \, ,  \label{meq2012} 
\end{align}
\begin{align}
	(1-\alpha) m & \lambda_n \, \mbox{bvec}  ({\bm Y}(\tilde{\bm \Omega}_{S^c})) 
	 = (1-\alpha) {\bm \Gamma}^\ast_{S^c,S} ({\bm \Gamma}^\ast_{S,S})^{-1} \nonumber \\
	  &  \times \Big( \mbox{bvec}(({\bm W}-{\bm R}( {\bm \Delta}))_S) 
		 + m \lambda_n \,  \mbox{bvec}({\bm Y}(\tilde{\bm \Omega}_S)) \Big) \nonumber \\
	& 	- (1-\alpha) \, \mbox{bvec}(({\bm W}-{\bm R}( {\bm \Delta}))_{S^c})
	 \, .  \label{meq2014}
\end{align}
A solution for ${\bm Z}(\tilde{\bm \Omega}_{S^c})$ is obtained via (\ref{meq2012}) and a solution for ${\bm Y}(\tilde{\bm \Omega}_{S^c})$ is obtained via (\ref{meq2014}). Using the notation as explained in Sec.\ \ref{TA2}, consider an edge $f \in S^c$ (implying $\|{\bm \Omega}^\ast_f\|_F = 0$) with $e_f$ denoting one of the corresponding $m^2$ edges in the corresponding enlarged graph (implying $|\Omega^\ast_{e_f}| = 0$). By (\ref{meq2012}), with ${\bm A}= {\bm \Gamma}^\ast_{e_f,S} ({\bm \Gamma}^\ast_{S,S})^{-1} \in {\mathbb R}^{1 \times (ms) }$, $s=|S|$, we have
\begin{align} 
  \lambda_n  &\, | Z(\tilde{\bm \Omega}_{e_f}) |
	 \le | {\bm A} \,  \mbox{bvec}({\bm W}_S) | + | {\bm A} \, \mbox{bvec}(({\bm R}( {\bm \Delta}))_S) | \nonumber \\
	 & \;\;\;   + \lambda_n \, | {\bm A} \,  \mbox{bvec}({\bm Z}(\tilde{\bm \Omega}_S)) |
		+  \, |{\bm W}_{e_f}| + |({\bm R}( {\bm \Delta}))_{e_f}|
	   \label{meq2016} \\
	& \; \le \| {\bm A} \|_1 \Big( \|{\bm W}\|_\infty + \| {\bm R}( {\bm \Delta}) \|_\infty
	  + \lambda_n \, \| {\bm Z}(\tilde{\bm \Omega}_S) \|_\infty \Big) \nonumber \\
	& \;\;\;   + \|{\bm W}\|_\infty + \| {\bm R}( {\bm \Delta}) \|_\infty  \label{meq2018} 
\end{align}
Using (\ref{meq210}) and the fact that $|Z(\tilde{\bm \Omega}_{e_g})| \le 1$ for any $e_g \in g \in S$, we have
\begin{align} 
  \lambda_n  &\, | Z(\tilde{\bm \Omega}_{e_f}) |
	 \le (2-\gamma ) \big( \|{\bm W}\|_\infty + \| {\bm R}( {\bm \Delta}) \|_\infty \big)
	     + \lambda_n \, (1-\gamma) \nonumber \\
	& \quad \le (2-\gamma ) \gamma \lambda_n/2 + \lambda_n \, (1-\gamma) \, . \label{meq2022}
\end{align}
Thus 
\begin{align} 
  | Z(\tilde{\bm \Omega}_{e_f}) | &
	 \le \gamma - \frac{\gamma^2}{2} + 1 - \gamma = 
	  1 -  \frac{\gamma^2}{2} < 1 \, ,\label{meq2024}
\end{align}
establishing that (\ref{meq2012}) holds for some $Z(\tilde{\bm \Omega}_{e_f})$ with $| Z(\tilde{\bm \Omega}_{e_f}) | < 1$ (strict feasibility) for any $e_f \in f \in S^c$. We now turn to (\ref{meq2014}) where we need to show $\|\mbox{vec}({\bm Y}(\tilde{\bm \Omega}_f))\|_2 < 1$, ${\bm Y}(\tilde{\bm \Omega}_f) \in {\mathbb R}^{m \times m}$, for any $f \in S^c$. By (\ref{meq2014}), with ${\bm B}= {\bm \Gamma}^\ast_{f,S} ({\bm \Gamma}^\ast_{S,S})^{-1} \in {\mathbb R}^{m \times (ms) }$, $s=|S|$, we have
\begin{align} 
 m \lambda_n  &\, \|\mbox{vec}({\bm Y}(\tilde{\bm \Omega}_f))\|_2
	 \le \| {\bm B} \,  \mbox{bvec}({\bm W}_S) \|_2  \nonumber \\
	  & \;\; + \| {\bm B} \, \mbox{bvec}(({\bm R}( {\bm \Delta}))_S) \|_2 
	    + m \lambda_n \, \| {\bm B} \,  \mbox{bvec}({\bm Y}(\tilde{\bm \Omega}_S)) \|_2 \nonumber \\
		& \;\; +  \, \|\mbox{vec}({\bm W}_f) \|_2 + \|\mbox{vec}(({\bm R}( {\bm \Delta}))_f) \|_2
	   \label{meq2026} \\
	&  \le \| \bm{\mathcal C}({\bm B}) \|_1 \Big( \|\bm{\mathcal C}({\bm W})\|_\infty 
	         + \| \bm{\mathcal C}({\bm R}( {\bm \Delta})) \|_\infty  \nonumber \\
	& \;\;\;  + m \lambda_n \, \| \bm{\mathcal C}({\bm Y}(\tilde{\bm \Omega}_S)) \|_\infty \Big) 
	 + \|\bm{\mathcal C}({\bm W})\|_\infty  \nonumber \\
		& \quad \;
	       + \| \bm{\mathcal C}({\bm R}( {\bm \Delta})) \|_\infty  \label{meq2028} 
\end{align}
where we used the fact that $\| {\bm B} \,  \mbox{bvec}({\bm W}_S) \|_2  \le \| \bm{\mathcal C}({\bm B}) \|_1  \|\bm{\mathcal C}({\bm W}_S)\|_\infty$ following \cite[Eqn.\ (80)]{Tugnait2024} (see also \cite[Lemma 13]{Kolar2014}). 
Using (\ref{meq200}), the fact that $\|\mbox{vec}({\bm Y}(\tilde{\bm \Omega}_g)) \|_2 \le 1$ for any $g \in S$, and the bounds on $\|\bm{\mathcal C}({\bm W})\|_\infty$ and $\| \bm{\mathcal C}({\bm R}( {\bm \Delta})) \|_\infty$, we have
\begin{align} 
  m \lambda_n  &\, \|\mbox{vec}({\bm Y}(\tilde{\bm \Omega}_f))\|_2
	 \le (2-\gamma ) \Big( \| \bm{\mathcal C}({\bm W}) \|_\infty 
	  \nonumber \\
	& \quad + \| \bm{\mathcal C}({\bm R}( {\bm \Delta})) \|_\infty \Big)  + m \lambda_n \, (1-\gamma) \nonumber \\
	& \le \, (2-\gamma ) \gamma m \lambda_n/2 + m \lambda_n \, (1-\gamma) \, . \label{meq2030}
\end{align}
Thus 
\begin{align} 
  \|\mbox{vec}({\bm Y}(\tilde{\bm \Omega}_f))\|_2 &
	 \le \gamma - \frac{\gamma^2}{2} + 1 - \gamma = 
	  1 -  \frac{\gamma^2}{2} < 1 \, ,\label{meq2032}
\end{align}
proving that for some ${\bm Y}(\tilde{\bm \Omega}_f)$ with $\|\mbox{vec}({\bm Y}(\tilde{\bm \Omega}_f))\|_2 = \|{\bm Y}(\tilde{\bm \Omega}_f) \|_F < 1$ for any $f \in S^c$, (\ref{meq2014}) holds. Satisfaction of (\ref{meq2012}) and (\ref{meq2014}) implies that of (\ref{meq2010}), and hence, that of (\ref{meqa220}) and  (\ref{meq1050}), yielding the desired result. $\quad \blacksquare$

{\it Lemma 5}. Suppose that 
\begin{align} 
  r := & 2 \kappa_{\Gamma^\ast} \big( \| \bm{\mathcal C}({\bm W}) \|_\infty + m \lambda_n \big) \nonumber \\
	   \le & \min \Big( \frac{1}{3 \kappa_{\Sigma^\ast} d_n} \, , 
		 \frac{1}{3 \kappa_{\Gamma^\ast} \kappa_{\Sigma^\ast}^3 d_n} \Big) \, . \label{meq2100}
\end{align}
Then  $\tilde{\bm \Omega} = {\bm \Omega}^\ast + {\bm \Delta}$ of Lemma 4 satisfies $\| \bm{\mathcal C}({\bm \Delta}) \|_\infty \le r$. $\quad \bullet$ \\
{\it Proof}. Define the closed ball
\begin{align} 
 {\cal B}(r) := & \big\{ {\bm \Delta}_S \, : \, \| \bm{\mathcal C}({\bm \Delta}_S) \|_\infty \le r \big\}
              \label{meq2102}
\end{align}
and the gradient mapping (\ref{maeqn210})
\begin{align} 
 G({\bm \Omega}) := & \hat{\bm \Sigma} - {\bm \Omega}^{-1}  + {\bm X}({\bm \Omega}) \, .
              \label{meq2104}
\end{align}
By construction $G(\tilde{\bm \Omega}_S) := \hat{\bm \Sigma}_S - ({\bm \Omega}^{-1})_S  + {\bm X}({\bm \Omega}_S)  = {\bm 0}$ and $\tilde{\bm \Omega}_{S^c} = {\bm 0}$. As in (\ref{meq1050}), 
\begin{align} 
   G({\bm \Omega}^\ast + {\bm \Delta}) = & ({\bm \Omega}^\ast)^{-1} {\bm \Delta} ({\bm \Omega}^\ast)^{-1}  
	 - {\bm R}( {\bm \Delta}) \nonumber \\
	& + {\bm W}+ {\bm X}({\bm \Omega}^\ast + {\bm \Delta})   \, . \label{meq2106} 
\end{align} 
Since $\tilde{\bm \Omega}_{S^c} = {\bm 0}$, we have ${\bm \Delta}_{S^c} = {\bm 0}$. Vectorizing and using decomposition as in (\ref{meq2000a}), we have
\begin{align} 
  \mbox{bvec}&( G({\bm \Omega}_S^\ast + {\bm \Delta}_S)) =  {\bm \Gamma}^\ast_{S,S} \mbox{bvec}({\bm \Delta}_S )  
	   \nonumber \\
	 & \quad  + \mbox{bvec}(({\bm W}-{\bm R}( {\bm \Delta}))_S) + \mbox{bvec}({\bm X}(\tilde{\bm \Omega}_S)) \, .  \label{meq2108} 
\end{align}
Define a mapping $F({\bm \Delta}_S)$ on ${\cal B}(r)$ as
\begin{align} 
  F&({\bm \Delta}_S) :=  -({\bm \Gamma}^\ast_{S,S})^{-1} \mbox{bvec}( G({\bm \Omega}_S^\ast + {\bm \Delta}_S))
	   + \mbox{bvec}({\bm \Delta}_S ) \nonumber \\
		 &  = -({\bm \Gamma}^\ast_{S,S})^{-1} \Big( \mbox{bvec}(({\bm W}-{\bm R}( {\bm \Delta}))_S) 
		  + \mbox{bvec}({\bm X}(\tilde{\bm \Omega}_S)) \Big) \, . \label{meq2110} 
\end{align}
The proof technique of \cite{Ravikumar2011} (see also \cite{Kolar2014}) is to show that $F({\cal B}(r)) \subseteq {\cal B}(r)$ (i.e., $F({\bm \Delta}_S)$ maps ${\bm \Delta}_S \in {\cal B}(r)$ to $F({\bm \Delta}_S) \in {\cal B}(r)$). Since the mapping $F$ is continuous and ${\cal B}(r)$ is compact, by the Brouwer's fixed point theorem \cite[Theorem 9.2]{Pata2019} $F({\cal B}(r)) \subseteq {\cal B}(r)$ implies that there exists a fixed point ${\bm \Delta}_S \in  {\cal B}(r)$  such that $F({\bm \Delta}_S) = \mbox{bvec}({\bm \Delta}_S)$, which, in turn, leads to $G({\bm \Omega}^\ast_S + {\bm \Delta}_S) = G(\tilde{\bm \Omega}_S) = {\bm 0}$ establishing that the fixed point is a constrained stationary point of $\bar{\cal L}({\bm \Omega})$ with $\| \bm{\mathcal C}({\bm \Delta}) \|_\infty \le r$ since ${\bm \Delta}_{S^c} = {\bm 0}$. It remains to show that $F({\cal B}(r)) \subseteq {\cal B}(r)$. By (\ref{meq2110}), in a manner similar to (\ref{meq2028}), we have
\begin{align} 
  & \| \bm{\mathcal C}(F({\bm \Delta}_S)) \|_\infty 
		 \le  \| \bm{\mathcal C}( {\bm \Gamma}^\ast_{S,S})^{-1} \|_{1,\infty}
		    \Big( \| \bm{\mathcal C}({\bm W}) \|_\infty  \nonumber \\
				& \quad\quad + \| \bm{\mathcal C}({\bm R}( {\bm \Delta})) \|_\infty 
		  + \| \bm{\mathcal C}({\bm X}(\tilde{\bm \Omega}_S)) \|_\infty \Big) \, . \label{meq2112} 
\end{align}
Since $|Z(\tilde{\bm \Omega}_{e_g})| \le 1$ for any $e_g \in g \in S$, $\|\mbox{vec}({\bm Z}(\tilde{\bm \Omega}_{g}))\|_2 \le m$ using the Cauchy-Schwarz inequality. Also, $\|\mbox{vec}({\bm Y}(\tilde{\bm \Omega}_g)) \|_2 \le 1$ for any $g \in S$. Using these two facts and (\ref{meq1060}), we have
\begin{align}  
	& \| \bm{\mathcal C}({\bm X}(\tilde{\bm \Omega}_S) ) \|_\infty \le  
	  \alpha \lambda_n \| \bm{\mathcal C}( {\bm Z}(\tilde{\bm \Omega}_S) )  \|_\infty  \nonumber \\
	& \quad\quad		+ (1-\alpha)m \lambda_n \| \bm{\mathcal C}( {\bm Y}(\tilde{\bm \Omega}_S) ) \|_\infty  \nonumber \\
	& \quad  \le  \alpha \lambda_n m + (1-\alpha)m \lambda_n \, = \, m \lambda_n \, .   \label{meq2114}  
\end{align}
Therefore,
\begin{align} 
   \| \bm{\mathcal C}(F({\bm \Delta}_S)) \|_\infty 
		 \le &  \kappa_{\Gamma^\ast} \big( \| \bm{\mathcal C}({\bm W}) \|_\infty + 
		 \| \bm{\mathcal C}({\bm R}( {\bm \Delta})) \|_\infty + m \lambda_n \big) \, . \label{meq2116} 
\end{align}
Since $r \le 1/(3 \kappa_{\Sigma^\ast} d_n)$ and $\| \bm{\mathcal C}({\bm \Delta}_S) \|_\infty \le r$, by Lemma 3, $\| \bm{\mathcal C}({\bm R}( {\bm \Delta})) \|_\infty \, \le \,
	(3d_n/2) \, \| \bm{\mathcal C}({\bm \Delta})\|_\infty^2 
	 {\kappa}_{\Sigma^\ast}^3$. Hence
\begin{align} 
   \kappa_{\Gamma^\ast} & \| \bm{\mathcal C}({\bm R}( {\bm \Delta})) \|_\infty
		 \le  3 \kappa_{\Gamma^\ast} d_n {\kappa}_{\Sigma^\ast}^3 r^2/2 
		\le  \frac{1}{r} \, \frac{r^2}{2} = \frac{r}{2} \, . \label{meq2118} 
\end{align}	
Thus
\begin{align} 
   \| \bm{\mathcal C}(F({\bm \Delta}_S)) \|_\infty 
		 \le &  \frac{r}{2} + \kappa_{\Gamma^\ast} \big( \| \bm{\mathcal C}({\bm W}) \|_\infty  
		 + m \lambda_n \big) \nonumber \\
		= & \frac{r}{2}  + \frac{r}{2}  = r \, . \label{meq2120} 
\end{align}
Therefore, $F({\bm \Delta}_S) \in {\cal B}(r)$, yielding the desired result. $\quad \blacksquare$

We now turn to the proof of Theorem 4. \\
{\it Proof of Theorem 4}. Here we first verify the conditions in Lemmas 3-5. Pick $\lambda_n$ as in (\ref{eqn350}). By Lemma 2, this choice ensures that $\| \bm{\mathcal C}({\bm W}) \|_\infty \le \gamma \lambda_n m/4$ and $\| {\bm W} \|_\infty \le \gamma \lambda_n /4$ (needed in Lemma 4) with probability $> 1-1/p_n^{\tau -2}$ provided the sample size $n > N_1 = 2 \ln(4m^2 p_n^\tau)$.  Now consider
\begin{align} 
  r = & 2 \kappa_{\Gamma^\ast}  \big( \| \bm{\mathcal C}({\bm W}) \|_\infty + m \lambda_n \big) 
	\le 2 \kappa_{\Gamma^\ast} \big( 1+\frac{\gamma}{4}\big)  m \lambda_n   \nonumber \\
		= & 2 \kappa_{\Gamma^\ast} \big( 1+\frac{4}{\gamma} \big) \tilde{C}_0 \sqrt{\ln(p_n) / n} \, . \label{meq2122} 
\end{align}
In Lemma 5 to satisfy (\ref{meq2100}), we pick $n > N_5$ which ensures that with probability $> 1-1/p_n^{\tau -2}$,
\begin{align} 
   2 & \kappa_{\Gamma^\ast} \big( 1+\frac{4}{\gamma} \big) \tilde{C}_0 \sqrt{\ln(p_n) / n}
	  \le \min \Big( \frac{1}{3 \kappa_{\Sigma^\ast} d_n} \, , 
		 \frac{1}{3 \kappa_{\Gamma^\ast} \kappa_{\Sigma^\ast}^3 d_n} \Big) \, . \label{meq2124} 
\end{align}
By Lemma 5, we have $\| \bm{\mathcal C}({\bm \Delta}) \|_\infty \le r \le 1/(3 \kappa_{\Sigma^\ast} d_n)$, which invoking Lemma 3 implies that $\| \bm{\mathcal C}({\bm R}( {\bm \Delta})) \|_\infty \le (3/2) d_n \kappa_{\Sigma^\ast}^3 \,  \| \bm{\mathcal C}({\bm \Delta}) \|_\infty^2 \le (3/2) d_n \kappa_{\Sigma^\ast}^3 \, r^2$. Therefore, we have 
\begin{align} 
   & \| \bm{\mathcal C}({\bm R}( {\bm \Delta})) \|_\infty \nonumber \\
	 & \quad \le \Big( 6 d_n \kappa_{\Sigma^\ast}^3  \,
		  \kappa_{\Gamma^\ast}^2 \big( 1+\frac{4}{\gamma} \big)^2 \tilde{C}_0 \sqrt{\ln(p_n) / n} \Big)  \gamma \lambda_n m/4
			   \, . \label{meq2126} 
\end{align} 
We pick $n > N_6$ so that $\| \bm{\mathcal C}({\bm R}( {\bm \Delta})) \|_\infty \le \gamma \lambda_n m/4$ with probability $> 1-1/p_n^{\tau -2}$. It remains to show that in Lemma 4, the condition $\| {\bm R}( {\bm \Delta}) \|_\infty  \le \gamma \lambda_n /4$ holds. To this end we impose an additional condition on (\ref{meq2122}) as
\begin{align} 
  r \le & 2 \kappa_{\Gamma^\ast} \big( 1+\frac{4}{\gamma} \big) \tilde{C}_0 \sqrt{\ln(p_n) / n} 
	  \le \frac{1}{3 \bar{\kappa}_{\Sigma^\ast} \bar{d}_n}  \, , \label{meq2128} 
\end{align}
i.e., modify the bounds on $r$ given in Lemma 5 as
\begin{align} 
  r \le &   \min \Big( \frac{1}{3 \kappa_{\Sigma^\ast} d_n} \, , 
		 \frac{1}{3 \kappa_{\Gamma^\ast} \kappa_{\Sigma^\ast}^3 d_n} \, ,
		 \frac{1}{3 \bar{\kappa}_{\Sigma^\ast} \bar{d}_n} \Big) \, . \label{meq2130}
\end{align}
By Lemma 5 and (\ref{meq2130}), we have $\| {\bm \Delta} \|_\infty \le \| \bm{\mathcal C}({\bm \Delta}) \|_\infty \le r \le 1/(3 \bar{\kappa}_{\Sigma^\ast} \bar{d}_n)$, which invoking Lemma 3 implies that $\| {\bm R}( {\bm \Delta}) \|_\infty \le (3/2) \bar{d}_n \bar{\kappa}_{\Sigma^\ast}^3 \,  \| {\bm \Delta} \|_\infty^2 \le (3/2) \bar{d}_n \bar{\kappa}_{\Sigma^\ast}^3 \, r^2$. Therefore, we have 
\begin{align} 
   & \| {\bm R}( {\bm \Delta}) \|_\infty \nonumber \\
	 & \quad \le \Big( 6 \bar{d}_n \bar{\kappa}_{\Sigma^\ast}^3  \,
		  \kappa_{\Gamma^\ast}^2 \big( 1+\frac{4}{\gamma} \big)^2 \, m \, \tilde{C}_0 \sqrt{\ln(p_n) / n} \Big)  
			\gamma \frac{\lambda_n}{4}
			   \, . \label{meq2132} 
\end{align} 
We pick $n > N_7$ so that $\| {\bm R}( {\bm \Delta}) \|_\infty \le \gamma \lambda_n /4$ with probability $> 1-1/p_n^{\tau -2}$. Thus, we have proved Theorem 4(i). Theorem 4(ii) follows from Lemma 4. To prove part (iii), consider
\begin{align} 
   & \| \bm{\mathcal C}(\hat{\bm \Omega} - {\bm \Omega}^\ast) \|_F  = 
	\sqrt{\sum_{\{ k, \ell \} \in S } \| \hat{\bm \Omega}^{(k,\ell)} - ({\bm \Omega}^\ast)^{(k,\ell)} \|_F^2 } \nonumber \\
	& \quad \le  \sqrt{s_n^\ast + p_n} \, \| \bm{\mathcal C}(\hat{\bm \Omega} - {\bm \Omega}^\ast) \|_\infty \, , \label{meq2140} 
\end{align}
where in the last step above we used the Cauchy-Schwarz inequality. Finally, to establish part (iv), note that parts (i)-(iii) hold with probability $> 1-1/p_n^{\tau-2}$ (with high probability (w.h.p.)). If $\min_{(k,\ell) \in S} \| ({\bm \Omega}^\ast)^{(k \ell)} \|_F \ge 2\| \bm{\mathcal C}(\hat{\bm \Omega} - {\bm \Omega}^\ast) \|_\infty$, 
\begin{align}
   & \|\bm{\mathcal C}(\hat{\bm \Omega} - {\bm \Omega}^\ast) \|_\infty = \|\bm{\mathcal C}((\hat{\bm \Omega} - {\bm \Omega}^\ast)_S) \|_\infty \nonumber \\
	& \quad \le (1/2) \min_{(k,\ell) \in S} \| ({\bm \Omega}^\ast)^{(k \ell)} \|_F. 
\end{align}
For any edge $\{ k , \ell\} = f \in S$, using the Cauchy-Schwarz inequality, we have
\begin{align}
     & (1/2) \min_{(k,\ell) \in S} \| ({\bm \Omega}^\ast)^{(k \ell)} \|_F 
	\ge \| (\hat{\bm \Omega} - {\bm \Omega}^\ast)_f \|_F \nonumber \\
	& \quad \ge \| {\bm \Omega}^\ast_f \|_F - \| \hat{\bm \Omega}_f \|_F,
\end{align}
therefore, 
\begin{align}
   &\| \hat{\bm \Omega}_f \|_F \ge \| {\bm \Omega}^\ast_f \|_F - (1/2) \min_{(k,\ell) \in S} 
	         \| ({\bm \Omega}^\ast)^{(k \ell)} \|_F \nonumber \\
	& \quad  \ge (1/2) \min_{(k,\ell) \in S} \| ({\bm \Omega}^\ast)^{(k \ell)} \|_F >0,
\end{align}
while $\hat{\bm \Omega}_{S^c} ={\bm 0}$ w.h.p. $\quad \blacksquare$

{\it Proof of Theorem 5}. We note that in terms of $\tilde{R}$ and $\tilde{r}_n$, Theorem 4 implies that
\begin{align}
   \| \bm{\mathcal C}(\hat{\bm \Omega} - {\bm \Omega}^\ast) \|_F & \le \tilde{R} \tilde{r}_n \, .
\end{align}
If $1/\beta_{\min} \le 0.99 \, \bar{\mu}$, then $\bm{\Omega}^\ast \in {\cal B}$ since $\| \bm{\Omega}^\ast \| \le 1/\beta_{\min}$ by the stated assumption $\beta_{\min}  \le \phi_{\min}(\bm{\Sigma}^\ast)$. Now we establish that $\hat{\bm{\Omega}}  \in {\cal B}$. To this end, as in the proof of Theorem 2, consider 
\begin{align}  
		   \| \hat{\bm{\Omega}} \| \le & \| \hat{\bm{\Omega}} - \bm{\Omega}^\ast \|
			  + \| \bm{\Omega}^\ast \|  \nonumber \\
				 \le & \| \hat{\bm{\Omega}} - \bm{\Omega}^\ast \|_F
			  + \| \bm{\Omega}^\ast \|  \nonumber \\
				= & \| \bm{\mathcal C}(\hat{\bm \Omega} - {\bm \Omega}^\ast) \|_F
			  + \| \bm{\Omega}^\ast \|  \nonumber \\
		 \le & \tilde{R} \tilde{r}_n + 1/\beta_{\min} \, .
		\label{naeq8394b} 
\end{align}
Therefore, $\hat{\bm{\Omega}}  \in {\cal B}$. Thus, both  $\hat{\bm{\Omega}}$ and ${\bm{\Omega}}^\ast$ are feasible. The desired result then follows from Theorem 4 and (local) strict convexity of $\bar{\cal L}({\bm \Omega})$ over ${\cal B}$ implied by Lemma 1. $\quad \blacksquare$

\bibliographystyle{unsrt} 

\end{document}